\documentclass[runningheads]{llncs}

\usepackage[mobile]{eccv}

\usepackage{eccvabbrv}

\usepackage{graphicx}
\usepackage{booktabs}
\usepackage{color}
\usepackage[accsupp]{axessibility}  %

\usepackage{multicol}
\usepackage{multirow}
\usepackage{array}
\usepackage{printlen}
\uselengthunit{cm}

\usepackage{hyperref}

\usepackage{orcidlink}
\usepackage{marvosym} %
\newcommand{\method}{EDUS}

\newcommand{\bx}{\mathbf{x}}

\newcommand{\bw}{\mathbf{w}}

\newcommand{\bP}{\mathbf{P}}
\newcommand{\bC}{\mathbf{C}}

\newcommand{\bff}{\mathbf{f}}
\newcommand{\bF}{\mathbf{F}}

\newcommand{\bc}{\mathbf{c}}

\newcommand{\bd}{\mathbf{d}}
\newcommand{\br}{\mathbf{r}}

\newcommand{\nR}{\mathbb{R}}

\newcommand{\cL}{\mathcal{L}}

\makeatletter
\DeclareRobustCommand\onedot{\futurelet\@let@token\@onedot}
\def\@onedot{\ifx\@let@token.\else.\null\fi\xspace}
\def\eg{e.g\onedot}

\makeatother

\renewcommand{\eqref}[1]{Eq.~\ref{#1}}

\newcommand{\boldparagraph}[1]{\noindent{\bf #1:} }

\definecolor{orange}{rgb}{1,0.5,0}

\newif\ifcomment
\commenttrue
\ifcomment
	\newcommand{\ag}[1]{ \noindent {\color{orange} {\bf Andreas:} {#1}} }
	\newcommand{\yl}[1]{ \noindent {\color{cyan} {\bf Yiyi:} {#1}} }

\else
	\newcommand{\ag}[1]{}
	\newcommand{\yl}[1]{}
\fi

\newcolumntype{P}[1]{>{\centering\arraybackslash}m{#1}}

\begin{document}

\title{Efficient Depth-Guided Urban View Synthesis}

\author{Sheng Miao\inst{1}$^\ast$ \and
Jiaxin Huang\inst{1}$^\ast$ \and
Dongfeng Bai\inst{2} \and Weichao Qiu\inst{2} \and Bingbing Liu\inst{2} \and Andreas Geiger\inst{3,4} \and  Yiyi Liao\inst{1}$\textsuperscript{\Letter}$  }
\authorrunning{S.Miao, J.Huang et al.}

\institute{$^1$Zhejiang University \quad $^2$Huawei Noah's Ark Lab\quad $^3$University of Tübingen \quad $^4$Tübingen AI Center\\
 \url{https://xdimlab.github.io/EDUS/}
}

\maketitle
\renewcommand{\thefootnote}{\fnsymbol{footnote}}
\footnotetext[1]{Equally contributed. 
         $\textsuperscript{\Letter}$ Corresponding author.}

\begin{abstract}
  Recent advances in implicit scene representation enable high-fidelity street view novel view synthesis. However, existing methods optimize a neural radiance field for each scene, relying heavily on dense training images and extensive computation resources. To mitigate this shortcoming, we introduce a new method called Efficient Depth-Guided Urban View Synthesis (EDUS) for fast feed-forward inference and efficient per-scene fine-tuning. Different from prior generalizable methods that infer geometry based on feature matching, EDUS leverages noisy predicted geometric priors as guidance to enable generalizable urban view synthesis from sparse input images. The geometric priors allow us to apply our generalizable model directly in the 3D space, gaining robustness across various sparsity levels. Through comprehensive experiments on the KITTI-360 and Waymo datasets, we demonstrate promising generalization abilities on novel street scenes. Moreover, our results indicate that EDUS achieves state-of-the-art performance in sparse view settings when combined with fast test-time optimization.
  
  \keywords{urban view synthesis, generalizable NeRF, sparse view}
\end{abstract}

\section{Introduction}
\label{sec:intro} 
Novel View Synthesis (NVS) of street scenes is a key issue in autonomous driving and robotics. Recently, Neural Radiance Fields (NeRF)~\cite{nerf} have evolved as a popular scene representation for NVS. Several works{\cite{snerf,urbanradiancefield,unisim,f2nerf,emernerf}} subsequently demonstrate promising NVS performance for street scenes.

However, these techniques are not applicable to sparse image settings, as they are typical for autonomous driving. When vehicles move at high speeds, a substantial portion of the content is captured from merely two or three viewpoints. This insufficient overlap between consecutive views significantly hinders the performance of existing methods. Furthermore, as the cameras are in forward motion, the parallax angle between frames is typically small.
This reduction in parallax angle further amplifies the uncertainty of reconstruction.

Several recent methods address the sparse view setting: One line of works {\cite{ds-nerf,sparsenerf,mixnerf}} exploits external geometric priors or regularization terms, such as depth maps, geometric continuity, and color cues, to enhance model performance. However, these approaches require per-scene optimization, which may take up to a few hours to converge on a single scene.

\begin{figure}[t]
    \centering
    \includegraphics[width=\linewidth]{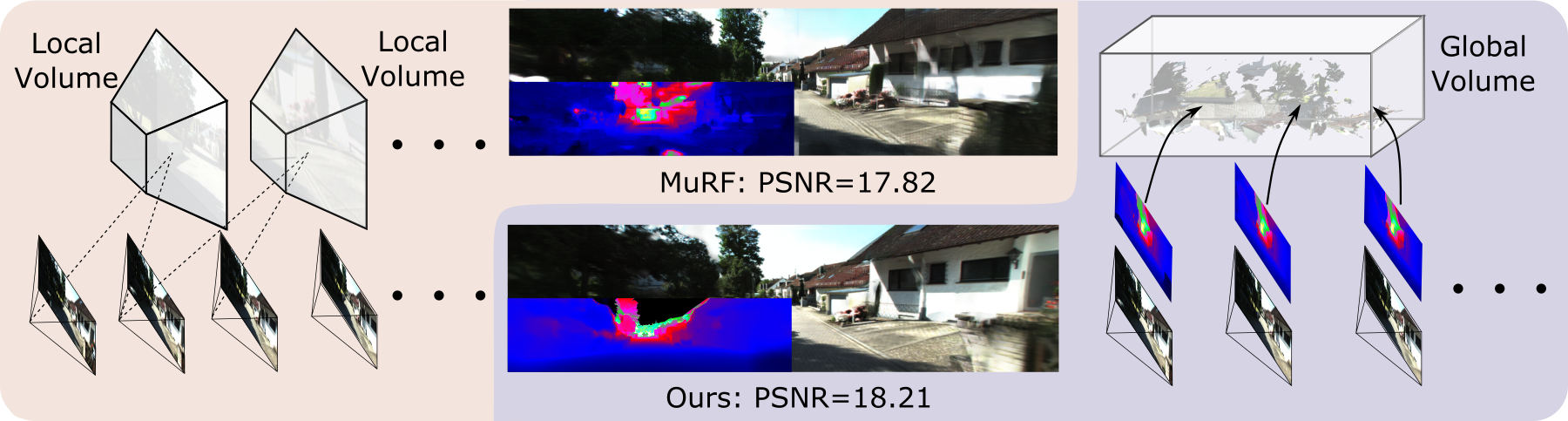}
    \caption{\textbf{Illustration}. Left: Most existing generalizable NeRF methods rely on feature matching for recovering the geometry, e.g., by constructing local cost volumes, potentially overfitting certain reference camera pose distributions. Right: Our method lifts geometric priors to the 3D space and fuses them into a global volume to be processed by a generalizable network. This enhances robustness as the geometric priors are unaffected by the reference image poses. We show synthesized images and depth maps through a feed-forward inference in the middle. }
    \label{fig:paralle_angle}
\end{figure}

Another line of research{\cite{ibrnet,pixelnerf,mvsnerf,neo360}}, known as generalizable NeRF, resorts to pre-training on large-scale datasets to gain domain-specific prior knowledge. In new scenarios, these data-driven methods are able to efficiently infer novel views from sparse inputs in a feed-forward manner, with the option for additional test-time optimization. Most of these methods rely on feature matching to learn the geometry, which suffers from insufficient overlap and textureless regions given sparse street views shown in \cref{fig:paralle_angle}. Furthermore, these methods typically retrieve the nearest reference images for feature matching, potentially leading to overfitting to specific camera pose configurations.
This hampers their generalization capability to unseen sparsity levels and various lengths of camera baselines, especially in highly sparse settings.
We hence ask the following question: \textit{Can we develop an efficient and generalizable urban view synthesis method capable of robustly handling diverse sparsity levels?}

To tackle this challenge, we propose Efficient Depth-Guided Urban View Synthesis (\method) which mitigates overfitting to a certain reference image pose configuration. This is achieved by lifting the reference images to 3D space utilizing approximate geometric priors, which are then processed by a feed-forward network directly applied in 3D space. 
In contrast to existing generalizable NeRF methods relying on feature matching for geometry prediction, whose performance is dependent on the relative poses between reference images and the target frame~\cite{mvsnerf,ibrnet}, our method explicitly leverages geometric priors, e.g., monocular depth. These priors remain unaffected by the reference image poses, making our method more robust to various sparsity levels. In addition, our method allows for accumulating geometric priors of reference frames within a large window (60 meters) to learn a generalizable model in this global volume. This further enables efficient per-scene fine-tuning by directly updating the global volume, circumventing the need to fine-tune numerous local volumes~\cite{mvsnerf} or the entire network~\cite{ibrnet, xu2023murf}. 

In order to leverage the geometric prior within a limited distance while being able to perform feed-forward inference for the full unbounded scene, we further decompose the scene into three components: foreground regions, background, and the sky, with a separate generalizable module designed for each.
Specifically, assuming a noisy point cloud of the foreground regions is provided by accumulating monocular/stereo depth estimations of the reference views, we train a generalizable 3D CNN to predict a feature volume. This 3D-to-3D generalizable module enhances robustness against various reference pose configurations as its input is the point cloud represented in world coordinates. 
While the 3D CNN leads to good geometry prediction, it encounters challenges in capturing high-frequency appearance details due to its inherent smoothness bias. To address this limitation, we augment the model with 2D features retrieved from nearby input views,
enhancing the model's ability to predict detailed appearance. 
Finally, we leverage image-based rendering for the background and sky. 

Our contributions in this paper can be summarized as follows: 1) We investigate the utility of noisy geometric priors and 3D-to-3D refinement for generalizable NeRF models, achieving promising feed-forward inference on various sparse urban settings. Our global volume-based representation further allows for efficient per-scene optimization which converges within five minutes.
2) We build our generalizable model tailored for unbounded street scenes by utilizing three separate generalizable components for the foreground, background, and the sky. This allows us to leverage the geometric priors within limited distances while being able to represent the full unbounded scene.
3) Our proposed approach outperforms baselines for sparse novel view synthesis on KITTI-360 dataset, including geometric regularization-based methods and generalization-based methods. We further demonstrate that our model trained on KITTI-360 shows convincing generalization performance on the Waymo dataset.

\section{Related Work}
\boldparagraph{Novel View Synthesis} Neural Radiance Fields~\cite{nerf} revolutionized the task of Novel View Synthesis (NVS).
Many follow-up works improve NeRF to achieve real-time rendering, fast training{\cite{sun2022direct, sun2022improved, yu2022plenoxels, liu2020neural}}, scene editing~\cite{haque2023instruct}, etc. In this work, we focus on generalizable NeRF on street views which has many applications in autonomous driving.

\boldparagraph{Street Scene NeRF}
There are a few NeRF-based methods carefully designed to deal with street scenes{\cite{urbanradiancefield, snerf, Neuralscenegraph, pnf}}.
One line of methods {\cite{urbanradiancefield,snerf,unisim}} combines LiDAR and RGB data for street scene reconstruction. In this work, we optionally leverage LiDAR in the generalizable training stage and consider an RGB-only setting for test scenes. Another line of methods consider the street scenes based on RGB images {\cite{Neuralscenegraph, pnf, wu2023mars, fu2022panoptic, fu2023panopticnerf}}%
. All aforementioned methods takes relatively dense images as input and consider per-scene optimization. Our method, in contrast, focuses on generalizable street scenes NeRF which is applicable to various sparsity levels.
Methods for unbounded scene representation are also relevant for street scenes {\cite{nerf-w, mipnerf360, f2nerf, barron2023zipnerf}}%
. Our method takes inspiration from these per-scene optimization methods and decomposes the scene into foreground, background, and sky with a generalizable module for each.

\boldparagraph{Sparse View NeRF}
One line of works leverages depth or appearance regularization {\cite{regnerf, ds-nerf, sparsenerf,mixnerf,jain2021putting,radford2021learning,freenerf}} to address the challenging task of sparse view synthesis. While leading to improvement in performance, these methods suffer from long training time.
In contrast, our method takes the geometric priors as input to train a generalizable NeRF, enabling feed-forward inference and fast per-scene optimization.
Generalizable NeRF approaches are also intensively studied in the literature~\cite{pixelnerf,ibrnet,mvsnerf, johari2022geonerf, liu2022neural, xu2023murf,xu2022point}.
Many works are built on image-based rendering~\cite{ibrnet,pixelnerf} or local cost volume {\cite{mvsnerf, johari2022geonerf, liu2022neural}}. As suggested in the current work MuRF~\cite{xu2023murf}, these methods struggle to generalize well to large baselines. MuRF tackles this issue by constructing a cost volume at the target view and achieves strong generalization to multiple baselines. However, it requires a long time for per-scene optimization, as the full computationally expensive model needs to be fine-tuned.
Our method is more closely related to PointNeRF{\cite{xu2022point}} and NeO 360{\cite{neo360}}, both reconstructing a scene representation in a global world coordinate.
PointNeRF leverages point clouds from a multi-view stereo algorithm and builds a point-based radiance field. We instead apply a 3D CNN to the input point cloud for refinement, allowing for better toleration of noisy point cloud input.
NeO 360{\cite{neo360}} proposes to learn a tri-plane-based global scene representation from sparse RGB images, enabling $360^{\circ}$ on the synthetic dataset. However, when it comes to complex street scenes with forward-moving cameras, NeO 360 struggles to directly predict the global scene representation. Our method tackles this challenge by incorporating explicit geometric priors.

\boldparagraph{NVS with Point Cloud} There are many existing NVS methods that take explicit geometric priors into account. {\cite{xu2022point, 3dgs}} learn a scene starting from a coarse point cloud, which can be obtained from multi-view stereo techniques{\cite{schonberger2016structure}} or learning-based methods{\cite{yao2018mvsnet, yin2023metric3d, yang2024depth}} by unprojecting depth maps. These point-based works combine explicit geometric priors and implicit networks to regress color and other attributes of each point in the point cloud, and then render 2D images through volume rendering or rasterization. Other methods {\cite{ruckert2022adop, li2023read}} project the point cloud into feature images first and then learn a model to fill holes during neural rendering. Instead, we only take as input the point cloud, extracting geometry and appearance guidance for the generalizable reconstruction. 

Some methods utilize encoders to process point clouds, followed by a volume-based neural rendering procedure. {\cite{huang2023ponder, zhu2023ponderv2}} mainly focus on pre-trained point cloud encoder for downstream tasks, and 
do not consider cases like sparse street views into consideration. {\cite{yang2023unipad}} reconstructs a driving scene well with either 2D images or 3D point cloud inputs, but it is limited by per-scene optimization. In contrast, our work is a generalizable method designed for autonomous driving scenes, which has a fast convergence speed and can obtain good reconstruction results even in extremely sparse viewpoints with fine-tuning.

\section{Method}
\begin{figure}[t]
    \centering
    \includegraphics[width=\linewidth]{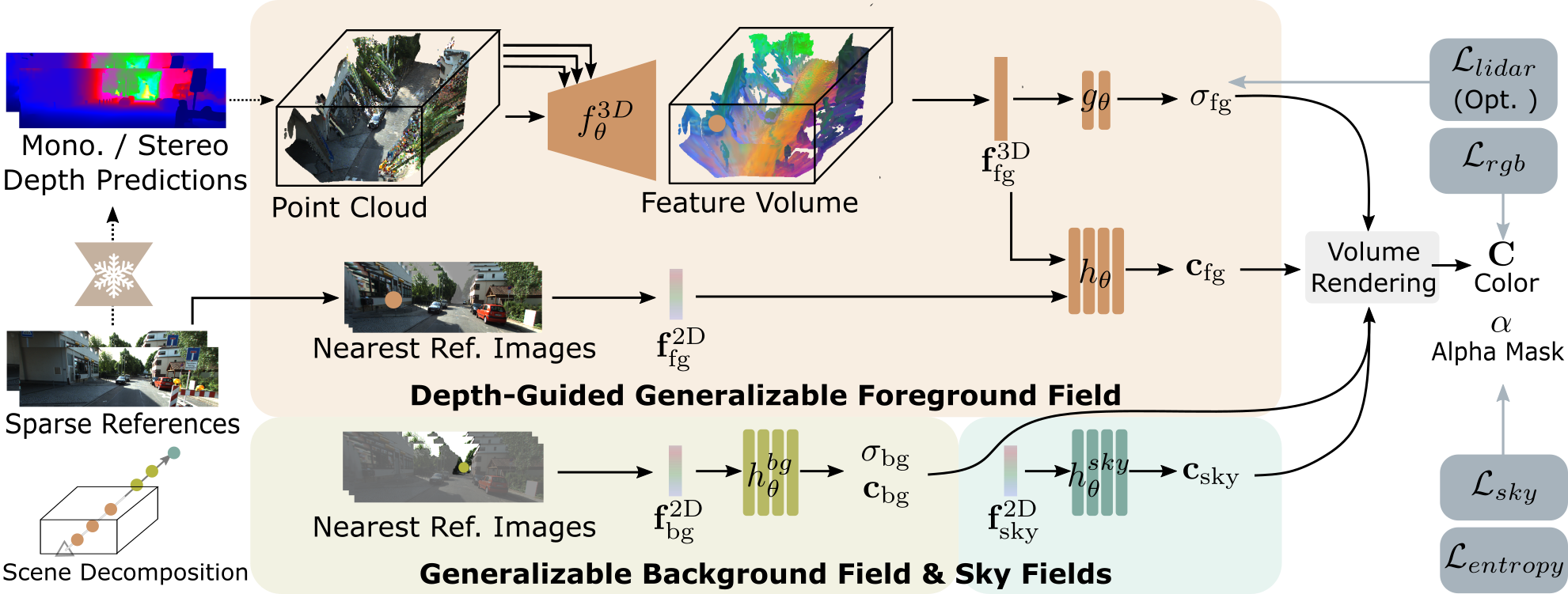}
    \caption{\textbf{\method}. Our model takes as input sparse reference images and renders a color image at a target view point. This is achieved by decomposing the scene into three generalizable modules: 1) depth-guided generalizable foreground fields to model the scene within a foreground volume; 2) generalizable background fields to model the background objects and stuff; and 3) generalizable sky fields. Our model is trained on multiple street scenes using RGB supervision $\cL_{rgb}$ and optionally LiDAR supervision $\cL_{lidar}$. We further apply  $\cL_{sky}$ to decompose the sky from the other regions and  $\cL_{entropy}$ to penalize semi-transparent reconstructions. }
    \label{fig:method}
\end{figure}

This paper explores the learning of a generalizable Neural Radiance Field (NeRF) for rendering novel views from sparse reference images in unbounded street scenes, with the option of test-time optimization. An overview of our \method ~is presented in \cref{fig:method}. 
Initially, we partition the scene into foreground, background, and sky components. This partitioning facilitates the characterization of foreground areas with restricted extent utilizing a 3D-to-3D adaptable framework, while the remaining distant regions are modeled through image rendering techniques. Throughout this paper, the term \textbf{foreground} denotes the region confined within a predetermined volume, while \textbf{background} encompasses substances and objects outside of this volume, excluding the \textbf{sky}. 
We develop distinct generalizable modules for each component. 
The core of our method is the depth-guided generalizable foreground field, effectively combining noisy geometric priors and image-based rendering for representing close-range regions (\cref{sec:pointcloud generation}). We further exploit image-based rendering to model the background regions and the sky (\cref{sec:bg_sky}). Finally, these three parts are composed to represent the unbounded street scenes (\cref{sec:decomposition}). Our method is trained on multiple street scenes and enables feed-forward NVS on unseen validation scenes, with optional fine-tuning for further improvement (\cref{sec:traing and finetuning}).

\subsection{Depth-Guided Generalizable Foreground Fields}
\label{sec:pointcloud generation}
\subsubsection{Depth Estimation:}
We explore geometric priors of existing powerful depth estimation methods to enhance our generalizable NeRF for foreground regions, i.e., regions within a close-range volume.
Given $N$ input images $\{I_{i}\}_{i=0}^{N}$ for a single scene, we exploit off-the-shelf depth estimators to predict metric depth maps $\{D_{i}\}_{i=0}^{N}$. 
The depth estimator can be a stereo depth model {\cite{shamsafar2022mobilestereonet}} when the input contains stereo pairs. Otherwise, we adopt a metric monocular depth estimator {\cite{yin2023metric3d}} which provides less accurate but plausible metric depth estimations.

\boldparagraph{Point Cloud Accumulation}
With the provided camera intrinsic matrices $\{K_{i}\}_{i=0}^{N}$ and camera poses $\{\mathbf{T}_{i}\}_{i=0}^{N}$, we unproject the predicted depth maps $\{D_{i}\}_{i=0}^{N}$ into 3D space, and accumulate them in the world coordinate system to form a scene point cloud $\mathcal{P} \in \mathbb{R}^{N_p \times 3}$.
Specifically, for each pixel with homogeneous coordinate $\mathbf{u}$ from a given frame, its world coordinate $\mathbf{x}$ is computed as: $\mathbf{x} = \left(d\mathbf{R}_i\mathbf{K}^{-1}\mathbf{u}+\mathbf{t}_i\right)$, 
where $d$ denotes the pixel's estimated depth and $\left(\mathbf{R}_i, \mathbf{t}_i\right)$ is derived from the $i$th frame's cam-to-world transformation $\mathbf{T}_{i}$. Additionally, we assign a 3-channel color value to each unprojected point by retrieving the color from the RGB images. Note that we keep points that fall within the foreground volume and omit the rest, as depth predictions are usually untrustworthy at far regions.

We also utilize depth consistency check to filter noise. Specifically, we unproject a depth map $\mathbf{D}_i$ of $i$th frame to 3D and reproject it to a nearby view $j$, obtaining a projected depth $\mathbf{D}_{i\rightarrow j}$. Next, we compare  $\mathbf{D}_{i\rightarrow j}$ and $\mathbf{D}_{j}$, and mask out pixels if the difference exceeds an empirical threshold $\sigma=0.2m$.

\boldparagraph{Modulation-based 3D Feature Extraction}
\label{sec:Construct Neural Volume}
The resulting noisy foreground point cloud serves as input for our generalizable feature extraction network $f_\theta^\text{3D}$. We first discretize the point cloud $\mathcal{P}$ into a volume $\bP \in \mathbb{R} ^ {H \times W \times D \times 3}$, where $H \times W \times D $ denotes the spatial resolution. The discretized point cloud is mapped to a feature volume $\bF\in\mathbb{R} ^ {H \times W \times D \times F}$, where $F$ denotes the feature dimension:
\begin{equation}
{\bF}=f_\theta^\text{3D}(\bP).
\end{equation}
Notably, the design choice of $f_\theta^\text{3D}$ affects the generalization performance. We experimentally observe that a conventional 3D U-Net with encoder-decoder architecture
results in blurry artifacts when generalizes to new scenes, as shown in supplementary material. %
Instead, we draw inspiration from \cite{yang2023urbangiraffe,spadecnn} to employ a 3D Spatially-Adaptive Normalization convolutional neural network (SPADE CNN) to modulate feature volume generation at multiple scales, which is a popular strategy in semantic image synthesis. More specifically, our SPADE CNN discards the encoder part and comprises 3 SPADE residual blocks and upsampling layers. For each block, SPADE CNN downsamples the input volume $\bP$ to the corresponding spatial resolution and uses it to modulate layer activations with learned scale and bias parameters. We hypothesize that SPADE CNN is more effective in preserving the appearance information encoded in $\bP$ thanks to its multi-resolution modulation. Given a sampled point $\bx\in \nR^3$ in the 3D space, we retrieve its feature $\bff_\text{bg}^\text{3D}\in\nR^{F}$ from $\bF$ using trilinear interpolation. More network details are provided in the supplementary materials.

\boldparagraph{Image-based 2D Feature Retrieval}
\label{sec:color retrive}
While the SPADE CNN takes as input the noisy depth predictions and achieves reasonable generalization, we observe that using $\bff_\text{bg}^\text{3D}$ alone struggles to provide high-frequency details in appearance. This is due to two reasons: 1) the point cloud discretization limits the spatial resolution; 2) the 3D CNN suffers from an inductive smoothness bias.
Therefore, we propose to additionally exploit the idea of image-based rendering to achieve better generalization of the appearance.
Inspired by \cite{ibrnet}, we first select $n$ nearest neighboring views from the target view based on the distances of the camera poses. This forms a group of nearest reference images $\hat{\mathcal{I}}=\{I_{k}\}_{k=0}^{K}$. For each sample point $\bx$, we project it to the reference frame in $\hat{\mathcal{I}}$. Next, we retrieve the colors from the reference frames based on bilinear interpolation, and concatenate them to form a 2D feature vector $\bff^\text{2D}_\text{fg}\in\nR^{3K}$. 
We set $K=3$ in this paper.

\boldparagraph{Color and Density Decoder}
\label{sec:decode radiance field}
Given a 3D point confined to the foreground volume, we predict its density $\sigma_\text{fg}$ and color $\bc_\text{fg}$ based on the volumetric feature $\bff_\text{fg}^\text{3D}$ and the 2D feature $\bff^\text{2D}_\text{fg}$. More specifically, the density $\sigma_\text{fg}$ is predicted by applying a tiny MLP $g_\theta$ to the volumetric feature $\bff_\text{fg}^\text{3D}$. 

For prediction of the color $\bc_\text{fg}$ , we concatenate $\bff_\text{fg}^\text{3D}$ and $\bff^\text{2D}_{fg}$ and feed them into a six-layer MLP. The 3D location $\bx$ and the viewing direction $\bd$ are further considered as input. This process can be expressed as:
\begin{equation}
\sigma_\text{fg}=g_\theta(\bff_\text{fg}^\text{3D}) , \quad \bc_\text{fg} =h_\theta(\bff_\text{fg}^\text{3D},\bff_\text{fg}^\text{2D},\gamma(\bx),\bd)
\label{eq:foreground}
\end{equation}
Different from the original NeRF \cite{nerf}, we apply positional encoding $\gamma(\cdot)$ to the location $\bx$ but not to the viewing direction $\bd$ as we expect non-Lambertion objects to be mainly modeled by the 2D feature $\bff_\text{fg}^\text{2D}$.
Please refer to the supplementary material for more details.

\subsection{Generalizable Background and Sky Fields}
\label{sec:bg_sky}
As mentioned above, the foreground volume only covers a limited street scene and falls short in representing the distant landscape (often hundreds of meters away) and the sky. Therefore, we model the background region and the sky separately.

\boldparagraph{Image-based Background Modelling}
An object placed outside the foreground volume is regarded as a background landscape. Due to perspective projection, distant objects occupy only a small portion of the image and have less detailed appearance. 
In this case, we observe that image-based rendering is sufficient to faithfully reconstruct the background regions as relative depth changes are small.
Thus, we employ an MLP $h_\theta^{bg}$ to jointly predict the background density $\sigma_\text{bg}$ and the color $\bc_\text{bg}$ based on 2D image-based feature $\bff_\text{bg}^\text{2D}$:

\begin{equation}
\sigma_\text{bg}, \bc_\text{bg} =h_\theta^{bg}(\bff_\text{bg}^\text{2D},\gamma(\bx),\bd)
\label{eq:background}
\end{equation}
where $\bx$ is spatial location and $\bd$ is the view direction. Same to the foreground, we retrieve color from nearby images and concatenate them to form vector $\bff_\text{bg}^\text{2D}$. 

\boldparagraph{Sky Modelling} 
Street scenes invariably contain the infinite sky region where rays do not collide with any physical objects, leading to minimal changes in appearance while moving forward.  Thus, we omit the positional influence and represent the sky as a view-dependent environment map. Given that the sky's appearance between consecutive frames is remarkably similar, we retrieve the 2D image feature $\mathbf{f}_{\text{sky}}^{\text{2D}}$ from reference frames, as we discussed in \cref{sec:color retrive}, and then blend the sky color using a single-layer MLP $h_{\theta}^{\text{sky}}$conditioned on view directions:
\begin{equation}
\mathbf{c}_{\text{sky}} = h_{\theta}^{\text{sky}}(\mathbf{f}_{\text{sky}}^{\text{2D}}, \mathbf{d}).
\end{equation}

\begin{figure}[t]
    \centering
    \setlength{\tabcolsep}{1pt}
    \def\mywidth{.33}
    \begin{tabular}{cccc}

    \includegraphics[width=\mywidth\linewidth]{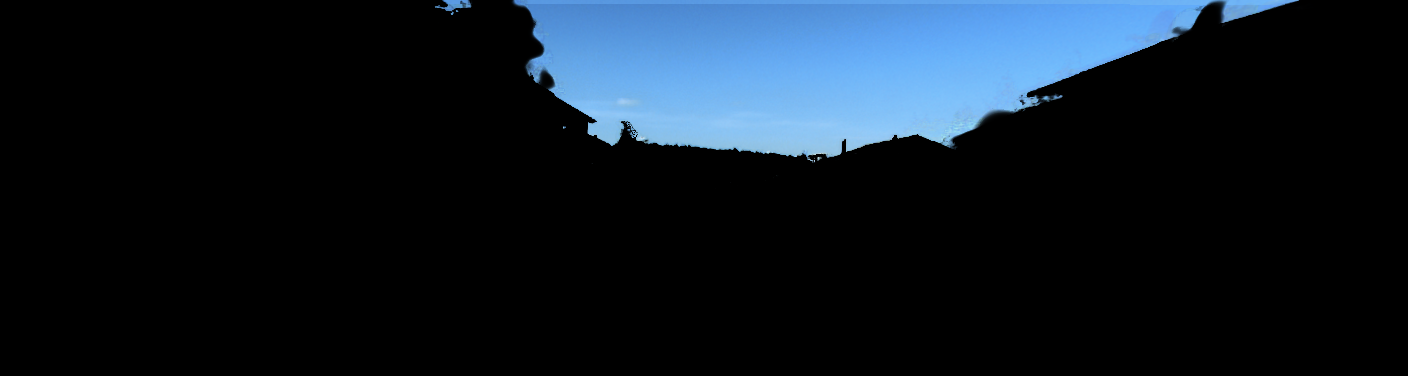} &
    \includegraphics[width=\mywidth\linewidth]{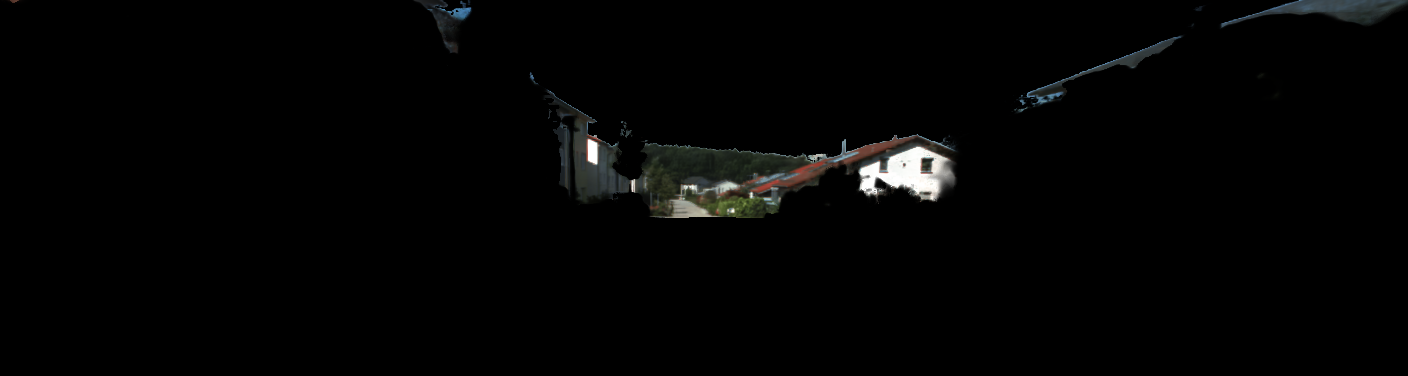} &
    \includegraphics[width=\mywidth\linewidth]{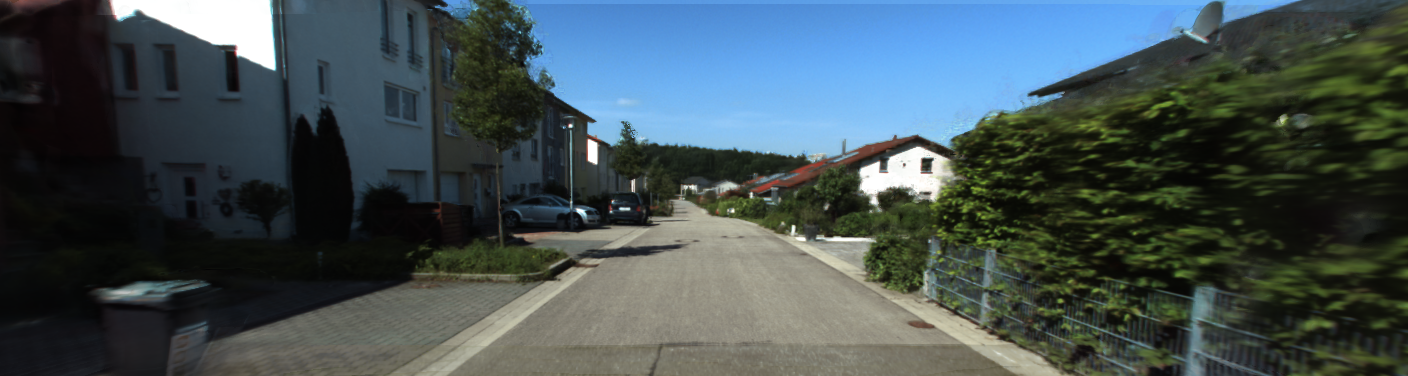} \\

    \begin{small}Sky~\end{small} &
    \begin{small}Background~\end{small} &
    \begin{small}Full Image\end{small} &
    \end{tabular}
    \caption{{\bf Scene Decomposition Visual Results}.}  
    \label{fig:composition}
\end{figure}

\subsection{Scene Decomposition}
\label{sec:decomposition}
The composition of the foreground, background, and sky leads to our full model for the unbounded street scenes, as shown in \cref{fig:composition}.
Here, we elaborate on the compositional volume rendering for each ray.
To render a pixel, we cast a ray $\br=\br_o + t\bd$ from the camera center $\br_o$ along its view direction $\bd$. We sample points along the ray to query their color and density. The color and density of  points falling within the foreground volume are queried through \cref{eq:foreground}, otherwise the background model in \cref{eq:background} is applied.

We accumulate all foreground and background points to obtain the color and the accumulated alpha mask:
\begin{equation}
 {\bC}^{(\mathrm{fg} + \mathrm{bg})}=\sum_{i=1}^{M} T_k\left(1-\exp \left(-\sigma_i\right)\right) \mathbf{c}_i, \quad
  {\alpha}^{(\mathrm{fg} + \mathrm{bg})}=\sum_{i=1}^{M} T_k\left(1-\exp \left(-\sigma_i\right)\right)
\end{equation}
where $T_i=\exp \left(-\sum_{j=1}^{i-1} \sigma_j\right)$ denotes the accumulated transmittance.
Finally, we add the sky color of the ray assuming it is located at an infinite distance:
\begin{equation}
 \bC = {\bC}^{(\mathrm{fg} + \mathrm{bg})} + (1-\alpha^{(\mathrm{fg} + \mathrm{bg})} )\bc_\text{sky}
\end{equation}

\subsection{Training and Fine-tuning}
\label{sec:traing and finetuning}
we use multiple training losses to train on different scenes and (optional) test-time optimization on a specific scene.
The most common sensors used in self-driving cars are cameras and LiDAR. Compared to cameras, LiDAR data are more expensive to acquire and may not be deployed in unseen test scenarios. As a result, we use LiDAR information only for training, to enhance geometric understanding of the model. This leads to the training loss function and fine-tuning loss functions as follows:
\begin{align}
\mathcal{L}_{\text {training }} & =\mathcal{L}_{\text {rgb}}+\lambda_1 \mathcal{L}_{\text {lidar}}+\lambda_2 \mathcal{L}_{\text {sky}}+\lambda_3 \mathcal{L}_{\text {entropy }} \\
\mathcal{L}_{\text {fine-tuning }} &=\mathcal{L}_{\text {rgb}}+\lambda_2 \mathcal{L}_{\text {sky}}+\lambda_3 \mathcal{L}_{\text {entropy }}
\end{align}
We now elaborate on the loss terms.

\boldparagraph{Photometric loss}
We use the $\mathcal{L}_2$ loss to minimize the photometric loss between the rendered colors ${\bC}$ and ground truth pixel color $\bC$:
\begin{equation}
\mathcal{L}_{\text {rgb}}=\left\|{\bC}-\bC_{gt}\right\|_2^2
\end{equation}

\boldparagraph{Sky loss} 
In order to separate the sky from the foreground and background solid structures, we leverage a pre-trained segmentation model \cite{tao2020hierarchical} to provide pseudo sky mask $M^{\text{sky}}$, and supervise the rendered sky mask with binary cross entropy (BCE) loss:
\begin{equation}
\mathcal{L}_{\text{sky}} = \text{BCE}\left(1-\alpha^{(\text{fg}+\text{bg})}, M^{\text{sky}}\right)
\end{equation}

\boldparagraph{LiDAR loss} During training, we optionally exploit LiDAR as an extra sensor to improve the reconstruction of our method. Given a set of collected lidar samples $\mathbf{r}(z)=\mathbf{o}_{\ell}+z \mathbf{d}_{\ell}$, Urban Radiance Fields {\cite{urbanradiancefield}} proposes line-of-sight loss which encourages the ray distribution to approximate the Gaussian Distribution $\mathcal{N}\left(0,\epsilon ^2\right)$, with $\epsilon$ being the bound width of the solid structure. 
As the assumption of Gaussian distribution may not hold for all rays, we experimentally observe that the loss in {\cite{urbanradiancefield}} leads to unstable training in our setting. 
Hence, we propose to soft the strict constraints as follows:

\begin{equation}
\mathcal{L}_{\text {empty }}=\int_{t_n}^{z-\epsilon} w(t)^2 d t,~ 
\mathcal{L}_{\text {near }}=1 - \int_{z-\epsilon}^{z+\epsilon}{w(t)} d t, ~
\mathcal{L}_{\text {dist }}=\int_{z+\epsilon}^{t_f} w(t)^2 d t
\end{equation}
where the bound width $\epsilon$ is set to 0.5 meters as initialization and we apply exponential decay to the minimum $0.1$ meters, which enables gradually decreasing ambiguities along the training step. Note that the second term $\mathcal{L}_{\text {near }}$ encourages our model to increase volumetric density within a certain range but without specifying its distribution,
whereas the first and third terms are required to keep the space empty for the rest of the regions.

\boldparagraph{Entropy regularization loss}
To penalize our model representing the distant landscape as semi-transparent, we introduce the entropy regularization loss to encourage opaque rendering followed by StreetSurf\cite{streetsurf}:
\begin{equation}
\mathcal{L}_{\text {entropy }}=-(\alpha^{\mathrm{fg}} \ln \alpha^{\mathrm{fg}}+(1-\alpha^{\mathrm{fg}}) \ln (1-\alpha^{\mathrm{fg}}))
\end{equation}

\section{Implementation details}
\boldparagraph{Input Volume Masking}
In the training stage, we randomly mask small regions of the input volume $\bP$. Similar to Masked AutoEncoders~\cite{mae}, this trick improves the completion capability when given a sparse and incomplete point cloud. More details and ablation results are shown in the supplementary material.

\boldparagraph{Hierarchical Sampling}
For rendering a ray, we follow NeuS~\cite{wang2021neus} to perform hierarchical sampling during training and inference. This is achieved by distributing the initial half of these samples uniformly, followed by iterative importance sampling for the rest. 

\boldparagraph{Appearance Embedding}
The images collected from urban scenarios frequently present variable illumination or other environmental changes. Inspired by \cite{nerf-w,blocknerf}, we add per-frame appearance embedding to account for exposure variations. For feed-forward inference on unseen scenes, we set the appearance embeddings as the mean embedding averaged across all training frames. Once fine-tuning is considered, we learn the appearance embedding of the reference images and interpolate them for the test views.

\boldparagraph{Training Details}
We train our neural network end-to-end using Adam optimizer\cite{kingma2014adam} with initial learning rate $5 \times10^{-3}$. We set $\lambda_1=0.1$, $\lambda_2=1$ and $\lambda_3=0.002$ as loss weights in our work. During training, we randomly select a single image and randomly sample 4096 pixels as a batch. Our \method~is trained on one RTX4090 GPU with 500k steps, which takes about 2 days.

\section{Experiment}

\boldparagraph{Dataset}
We conduct generalizable training on the KITTI-360 dataset~\cite{liao2022kitti}. 
Specifically, we gather 80 sequences encompassing varied street perspectives for training purposes, with each sequence comprising 60 stereo images alongside their respective camera poses. In the case of training scenes, we optionally utilize accumulated LiDAR points to supervise the geometry. Across all scenes, we standardize the pose by centering the translation around the origin and establish the axis-aligned bounding box (AABB) to decompose the street scene. The foreground range is set to $[-12.6,12.6 \mathrm{~m}]$ for X axes, $[-3,9.8 \mathrm{~m}]$ for Y axes, $[-20,31.2 \mathrm{~m}]$ for Z axes, and dimension of the voxelized pointcloud is $128 \times 64 \times 256$ since the voxel size is set to $(0.2 \mathrm{~m}, 0.2 \mathrm{~m}, 0.2 \mathrm{~m})$.

We use five public validation sets from the KITTI-360 dataset and five scenes sourced from Waymo\cite{waymo} as our test scenarios, with the Waymo dataset serving as an out-of-domain test set. It is important to note that the test sets from KITTI-360 have no overlap with the training scenes. Each test sequence comprises 60 images covering approximately 60 meters. To further assess our robustness against various levels of sparsity, we consider three sparsity settings: 50\%, 80\%, and 90\% drop rates, where a higher drop rate indicates a more sparse set of reference images. To ensure fairness in comparison, we employ the same test images across all drop rates.

\boldparagraph{Depth} In this paper, we use both monocular and binocular depth to conduct our experiment. For test sequences in KITTI-360, depth used as both supervision and input is stereo depth generated by MobileStereoNet\cite{shamsafar2022mobilestereonet}. We only supervise the regularization-based methods by left-eye depth. For the geometry-guided methods, we use the left-eye depth values of corresponding training images and aggregate them into a point cloud after depth consistency check~\cite{cheng2023uc}. For experiments on Waymo, we employ Metric3D\cite{yin2023metric3d} to get metric monocular depth maps and unproject them to form point clouds for evaluation.

\boldparagraph{Baselines}
We conduct a comparative analysis of our method against two lines of approaches. Firstly, we compare our model with generalizable approaches: IBRNet~\cite{ibrnet}, MVSNeRF~\cite{mvsnerf}, NeO 360~\cite{neo360} and MuRF~\cite{xu2023murf}. We train each model from scratch on the 80 training sequences of KITTI-360, and assess their performance both with and without per-scene fine-tuning. For all generalizable methods, we use three reference frames for training and inference. Secondly, we compare with test-time optimization methods, including methods for sparse view settings: MixNeRF~\cite{mixnerf}, SparseNeRF~\cite{sparsenerf}, DS-NeRF~\cite{ds-nerf}, and one taking point clouds as input: 3DGS~\cite{3dgs}. To maintain parity in our evaluation, we furnish the depth maps utilized in our method to train depth-supervision approaches~\cite{sparsenerf,ds-nerf}, and employ our accumulated point cloud as the initialization for 3DGS~\cite{3dgs}.

\begin{table}[tb]
  \renewcommand{\arraystretch}{1.0}
  \centering
   \resizebox{!}{2.4cm}{
  \begin{tabular}{m{1.5cm} m{1.5cm}<{\centering} m{0.95cm}<{\centering} m{0.95cm}<{\centering} m{0.95cm}<{\centering} m{0.15cm} m{0.95cm}<{\centering} m{0.95cm}<{\centering} m{0.95cm}<{\centering} m{0.15cm} m{0.95cm}<{\centering} m{0.95cm}<{\centering} m{0.95cm}<{\centering}}
    \toprule
    
    \multirow{2}{*}{Methods} & \multirow{2}{*}{Setting} & \multicolumn{3}{c}{KITTI$360_{drop50\%}$} & & \multicolumn{3}{c}{KITTI$360_{drop80\%}$} & & \multicolumn{3}{c}{Waymo$_{drop50\%}$} \\ \cline{3-5}  \cline{7-9}  \cline{11-13}
              &           & PSNR$\uparrow$  & SSIM$\uparrow$  & LPIPS$\downarrow$ & & PSNR$\uparrow$  & SSIM$\uparrow$  & LPIPS$\downarrow$ & & PSNR$\uparrow$  & SSIM$\uparrow$  & LPIPS$\downarrow$ \\
    \midrule
    IBR-Net  & \multirow{5}{*}{\begin{minipage}{1.5cm} \centering No \\ per-scene \\ opt. \end{minipage}}  & 19.99 & 0.624 & 0.217 & & 15.96 & 0.469 & 0.354 & & 21.28 & \textbf{0.777} & 0.199 \\
    MVSNeRF &  & 17.73 & 0.618 & 0.328 & & 16.50 & 0.577 & 0.365 & & 19.58 & 0.662 & 0.278\\
    Neo360 &   & 13.73 & 0.394 & 0.624 & & 12.98 & 0.357 & 0.659 & & 14.07 & 0.541 & 0.708  \\
    MuRF &  & \textbf{22.19} & 0.741 & 0.264 & & 18.69 & 0.639 & 0.353 & & 23.12 & 0.779 & 0.318  \\
    \textbf{Ours} &  & 21.93  & \textbf{0.745}  & \textbf{0.178} & & \textbf{19.63} & \textbf{0.668} & \textbf{0.244} & & \textbf{23.16} & 0.761 & \textbf{0.189} \\[0.1cm] 
    \cline{1-13} 
     IBR-Net & \multirow{5}{*}{\begin{minipage}{1.5cm} \centering Per-scene \\ opt. \end{minipage}} & 21.17 & 0.657 & 0.199 & &  17.98 & 0.529 & 0.279 & & 23.39 & 0.825 & 0.163 \\
    MVSNeRF &   & 19.47 & 0.647 & 0.310 & & 18.06 & 0.602 & 0.353 & & 24.28 & 0.759 & 0.207 \\
    Neo360 &   & 17.92 & 0.489 & 0.566 & & 17.51 & 0.445 & 0.581 & & 22.59 & 0.670 & 0.522 \\
    MuRF &  & 23.71 & 0.762 & 0.233 & & 19.70 & 0.666 & 0.321 & & 28.30 & 0.846 & 0.175 \\ 
    \textbf{Ours} &   & \textbf{24.43}  & \textbf{0.793} & \textbf{0.136} && \textbf{20.91} & \textbf{0.712} & \textbf{0.220} && \textbf{28.45}&\textbf{0.834} &\textbf{0.132} \\
  \bottomrule
  \end{tabular}
  }
  \caption{\textbf{Quantitative Comparison} on five test scenes among generalizable methods. All models are trained on the KITTI-360 dataset using drop 50\% sparsity level.
  }
\label{tab:generalizable}
\end{table}

\begin{figure}[t]
    \centering
    \setlength{\tabcolsep}{1pt}
    \def\mywidth{1.0}
    \def\mywidthP{3.8cm}
    \begin{tabular}{P{0.4cm}P{\mywidthP}P{\mywidthP}P{\mywidthP}}
    \rotatebox{90}{IBRNet} & 
    \includegraphics[width=\mywidth\linewidth]{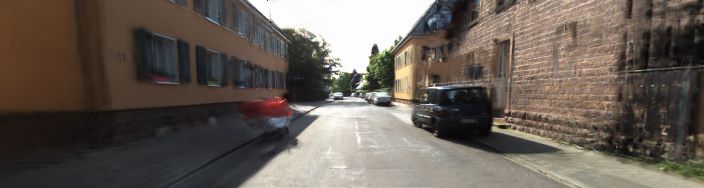}&
    \includegraphics[width=\mywidth\linewidth]{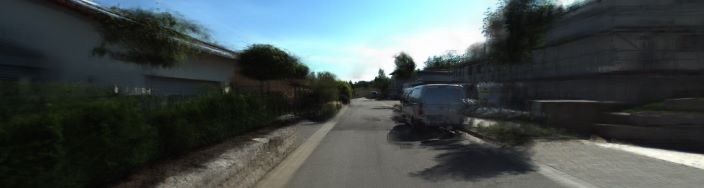}&
    \includegraphics[width=\mywidth\linewidth]{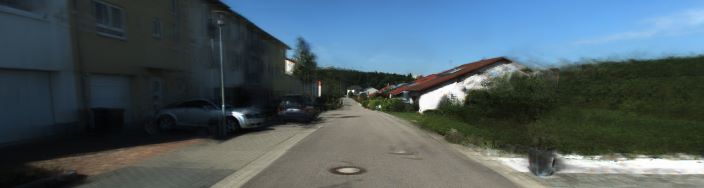}
    \\
    \rotatebox{90}{MuRF} & 
    \includegraphics[width=\mywidth\linewidth]{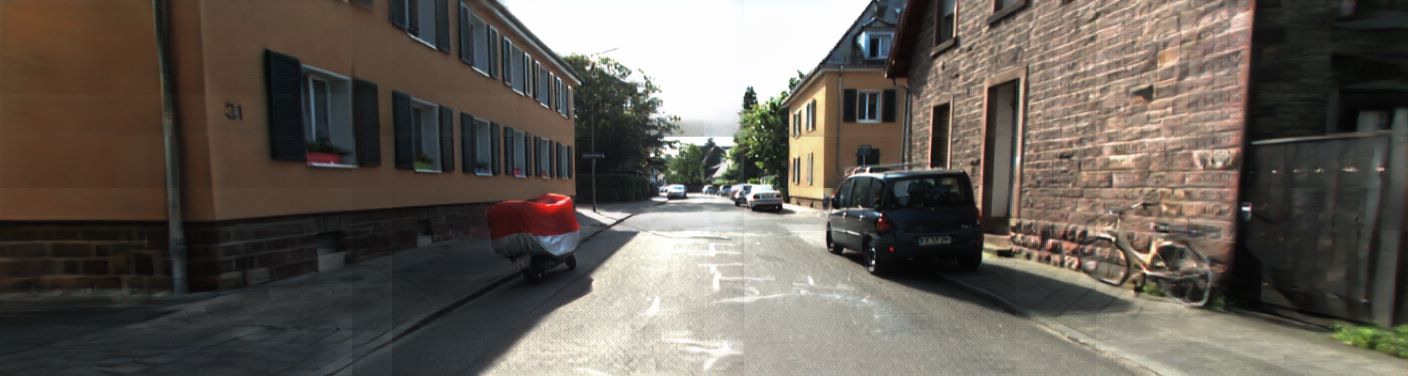}&
    \includegraphics[width=\mywidth\linewidth]{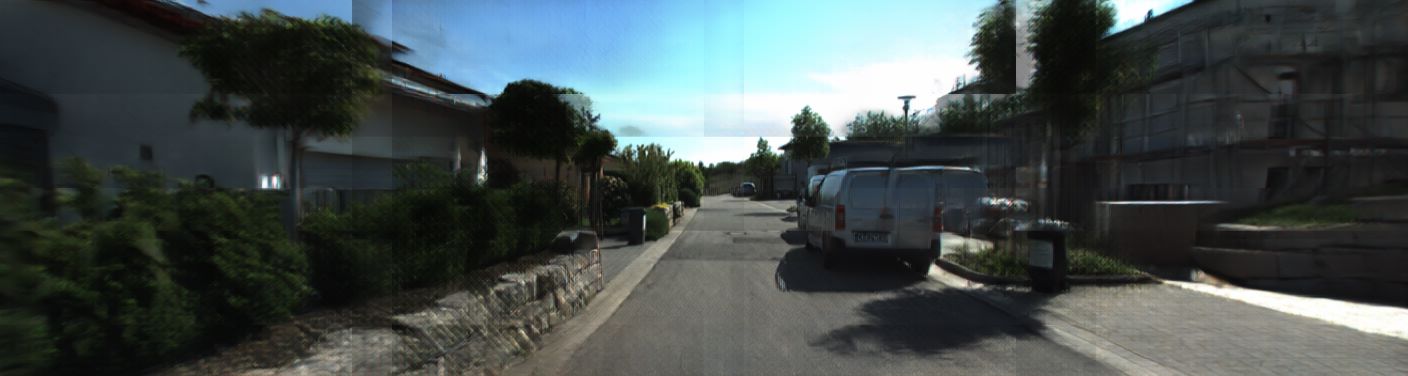}&
    \includegraphics[width=\mywidth\linewidth]{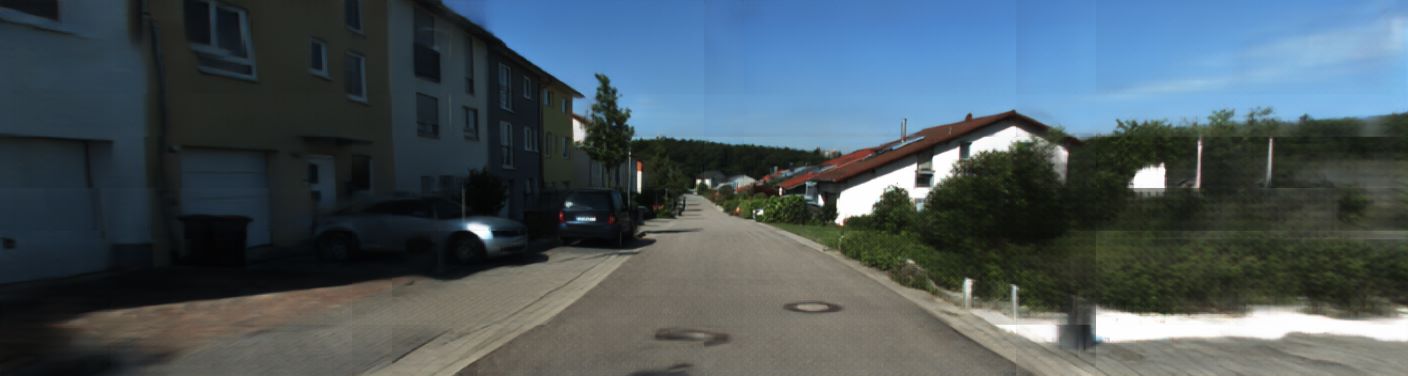} \\
    \rotatebox{90}{Ours} & 
    \includegraphics[width=\mywidth\linewidth]{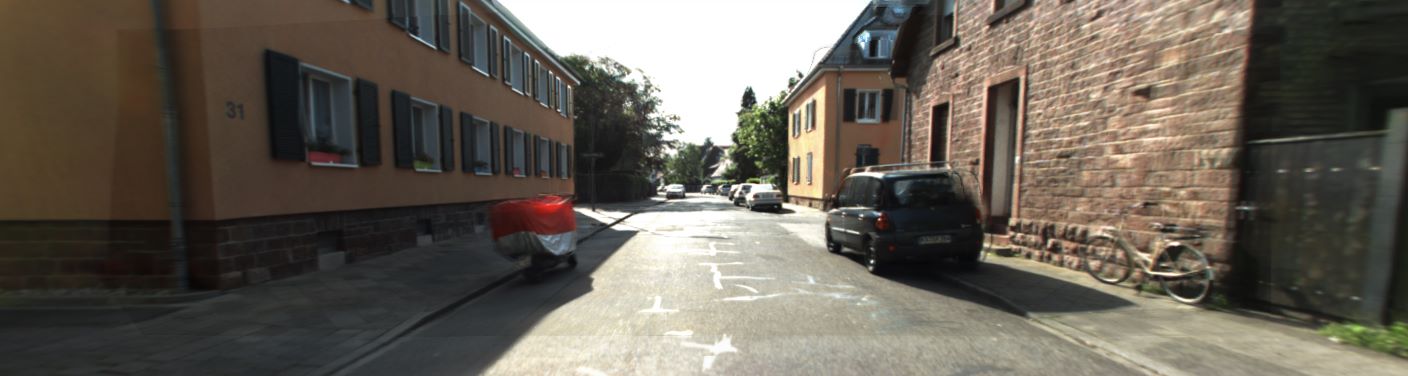}&
    \includegraphics[width=\mywidth\linewidth]{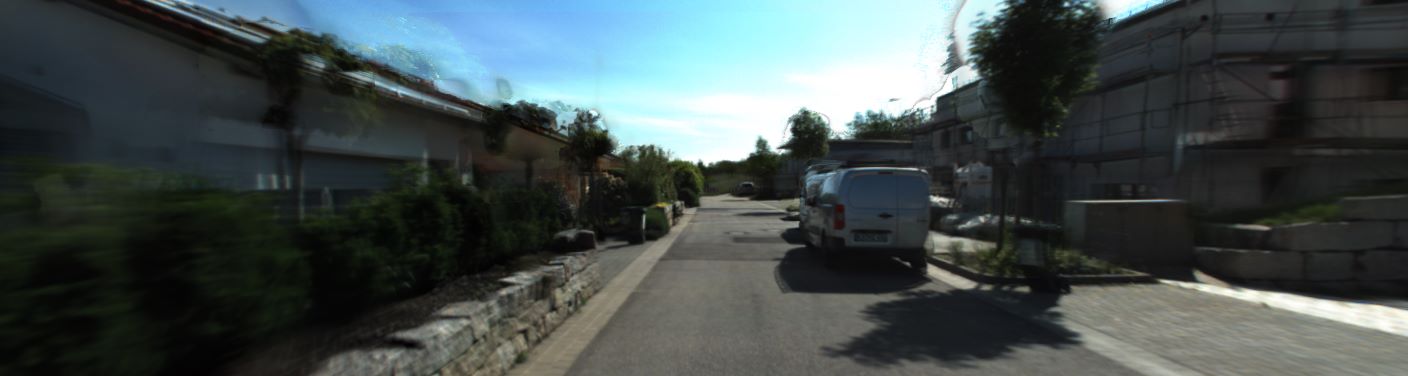}&
    \includegraphics[width=\mywidth\linewidth]{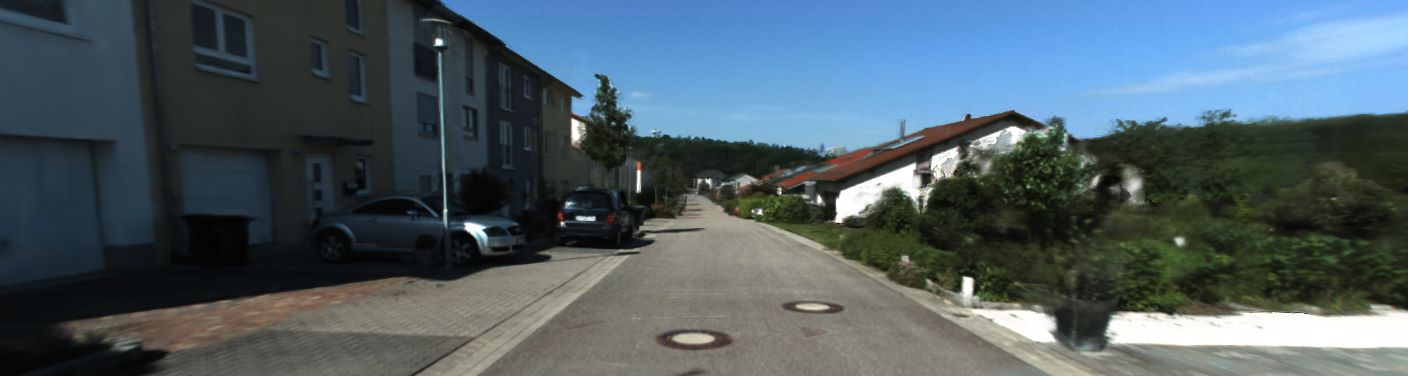} \\
    \rotatebox{90}{GT} & 
    \includegraphics[width=\mywidth\linewidth]{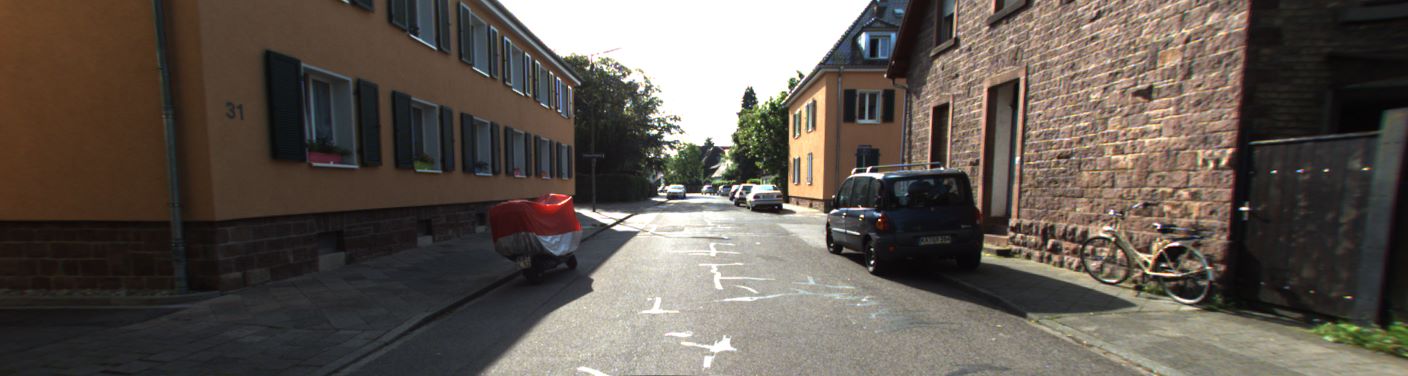}&
    \includegraphics[width=\mywidth\linewidth]{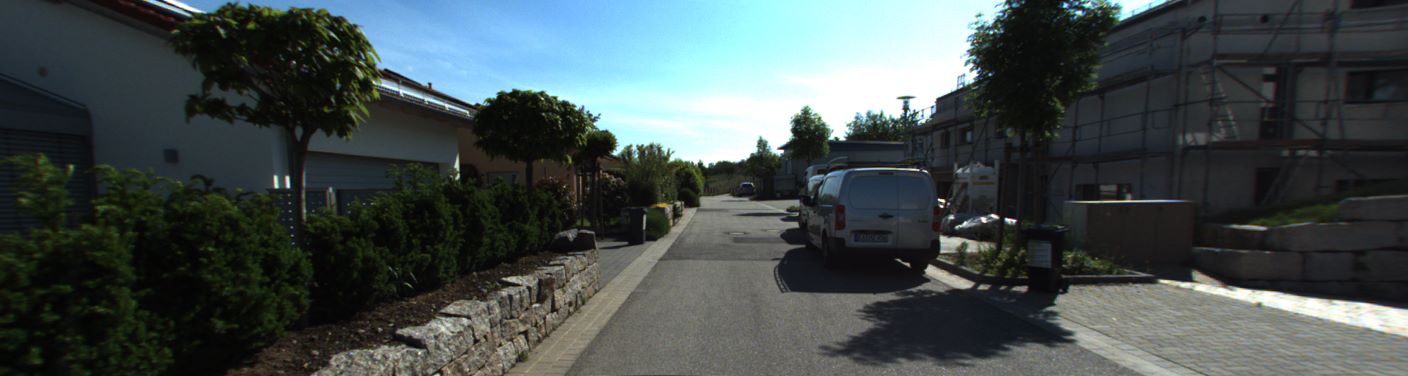}&
    \includegraphics[width=\mywidth\linewidth]{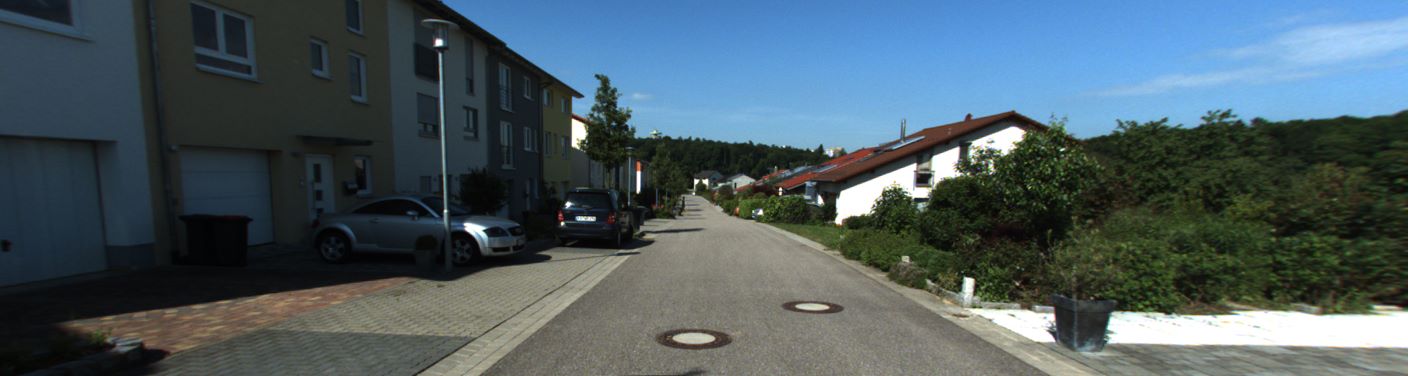} \\
    &
    \begin{small}Drop 50\% feed-forward~\end{small} &
    \begin{small}Drop 80\% feed-forward\end{small} &
    \begin{small}Drop 80\% fine-tuned\end{small} \\
    \end{tabular}
    \caption{{\bf Qualitative Comparison} with generalizable baselines.}  
    \label{fig:feed forward}
\end{figure}

\subsection{Comparison with Generalizable NeRFs}
\begin{figure}[t]
    \begin{minipage}{0.63\textwidth}
        \centering
        \vspace{0.23cm}
        \captionsetup{skip=10pt}
        \setlength{\tabcolsep}{2pt}
        \def\mywidth{.43}
        \def\myheight{.30}
        \begin{tabular}{cc}
        \includegraphics[height=\myheight\linewidth]{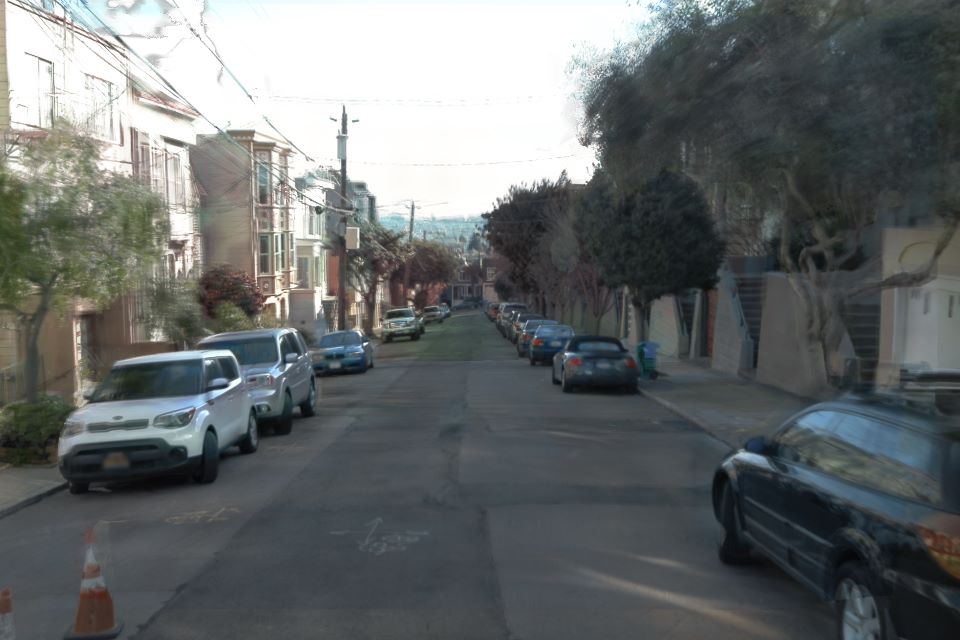}  &
        \includegraphics[height=\myheight\linewidth] {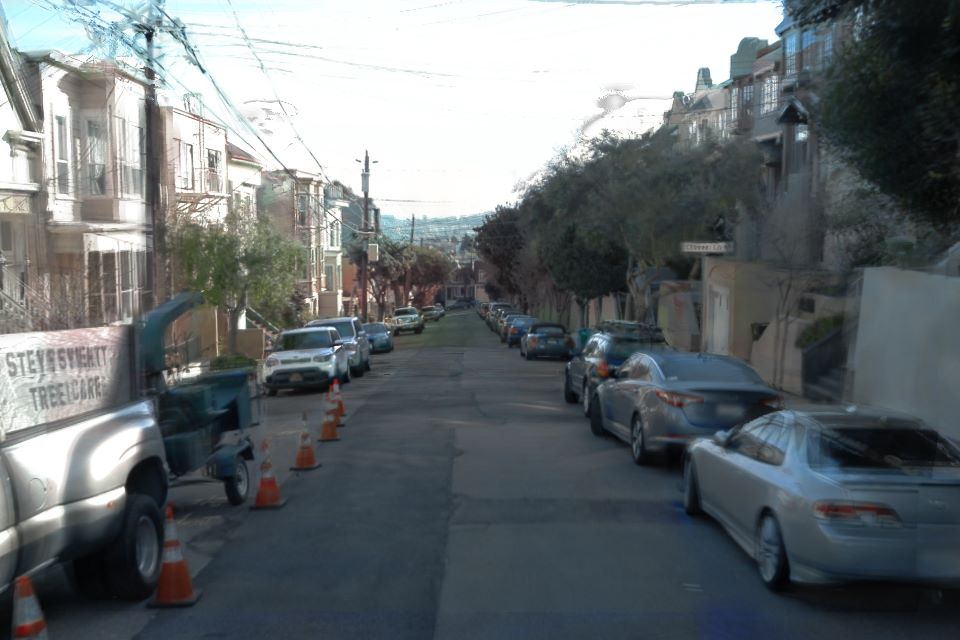} 
        \end{tabular}
        \caption{{\bf Zero-shot Inference on Waymo.} }
        \label{fig: zeroshot waymo}
    \end{minipage}
    \begin{minipage}{0.35\textwidth}
        \centering
        \captionsetup{skip=2pt}
        \def\timetable{.9}
        \includegraphics[width=\timetable\linewidth]{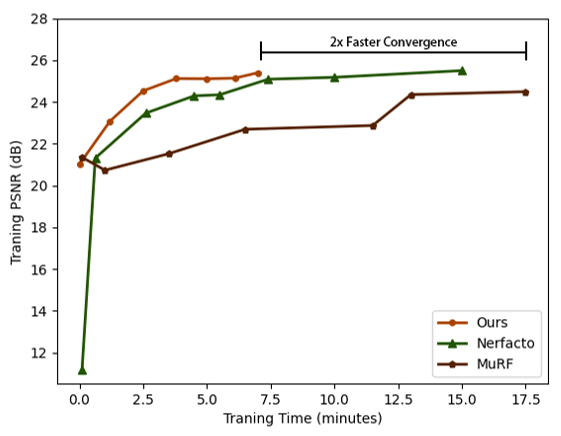}
        \caption{{\bf Training Time.} }
        \label{fig: training time}
    \end{minipage}
\end{figure}

\boldparagraph{Feed-Foward Inference}
We train all generalizable methods under 50\% drop rate setting on KITTI-360  and assess their performance using both the 50\% and 80\% drop rate sparsity levels on KITTI-360. \cref{tab:generalizable} and \cref{fig:feed forward} show the qualitative and quantitative comparisons respectively. The results show that our proposed \method~achieves photorealistic novel view synthesis on the complex street view using only feed-forward inference. 
More importantly, our method outperforms the baselines with a larger gap when evaluated at a higher unseen sparsity level (drop 80\%), as our method is less sensitive to the camera pose distribution. IBR-Net and MVSNeRF are mainly suitable for small-scale, object-level scenes and have difficulty handling unbounded street views, resulting in blurry images. 
NeO 360 directly predicts a tri-plane-based radiance field from RGB images, making it struggle to reconstruct the street scene of 60 meters driving distance. Our concurrent work MuRF \cite{xu2023murf} achieves comparable PSNR and SSIM in drop 50\% but sometimes shows blocky artifacts, resulting in worse LPIPS.  Furthermore, our methods outperform all baselines on the unseen Waymo datasets, demonstrating the generalization capability empowered by the geometric priors, see \cref{fig: zeroshot waymo}.

\boldparagraph{Per-Scene Fine-Tuning}
We further fine-tune all generalizable methods under different sparsity levels, as shown in \cref{tab:generalizable}. For each novel street sequence, we freeze our 3D CNN and directly fine-tune the feature volume to enable efficient convergence. Note that our feature volume is initialized by running a single feed-forward pass of the 3D CNN, yielding a good initialization and subsequently yielding improved performance than baselines after fine-tuning. 
 
Note that MuRF constructs local volume at each target view and requires complete propagation of gradients, resulting in slower convergence during per-scene optimization. In contrast, our global volume-based approach demonstrates faster convergence speed, see \cref{fig: training time}, making it more suitable for urban scenes. Besides, \method~shows superior memory efficiency when handling full-resolution ($376 \times 1408$) inference with 6GB while MuRF requires splitting the same size image into 4 patches, each consuming 16.2GB on a single RTX4090 (24GB).

\subsection{Comparison with Test-Time Optimization NeRFs}
\begin{figure}[tb]
    \centering
    \def\mywidth{.33}
    \captionsetup{skip=2pt} 
    \begin{tabular}{ccc}
      \includegraphics[width=\mywidth\linewidth]{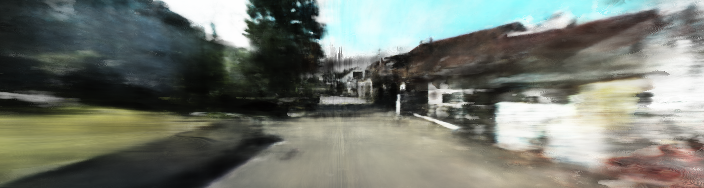}
      &
      \includegraphics[width=\mywidth\linewidth]{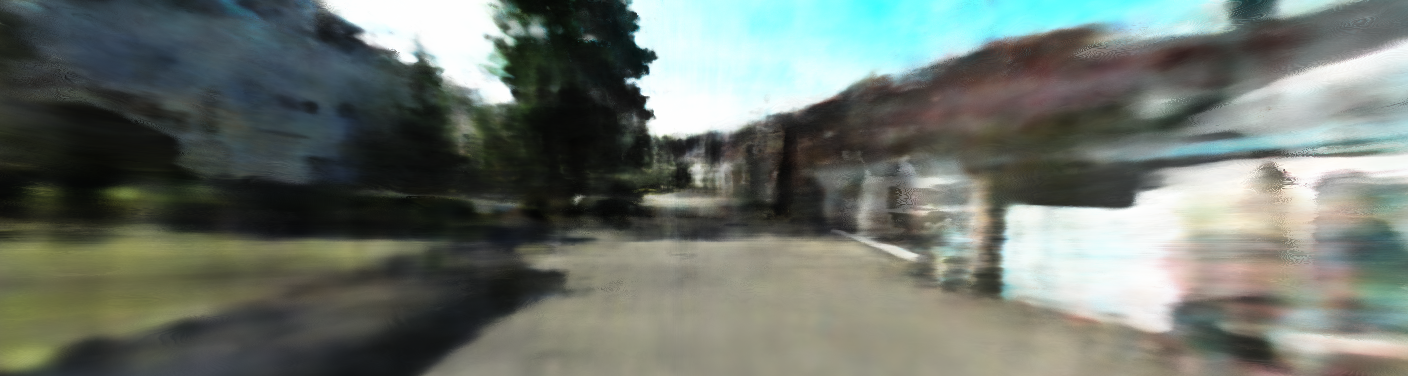}
      &
      \includegraphics[width=\mywidth\linewidth]{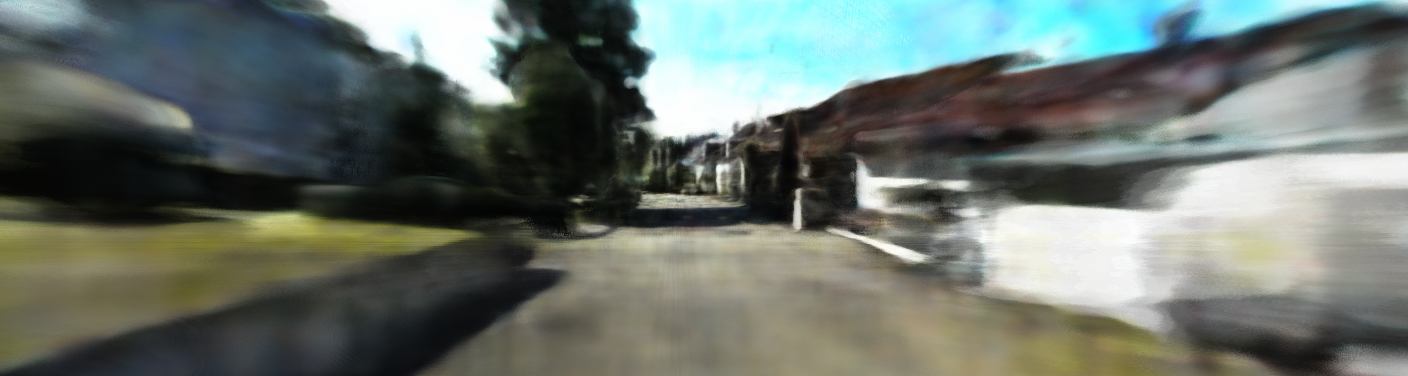}
      \\
      MixNeRF & SparseNeRF & DS-NeRF \\
      \includegraphics[width=\mywidth\linewidth]{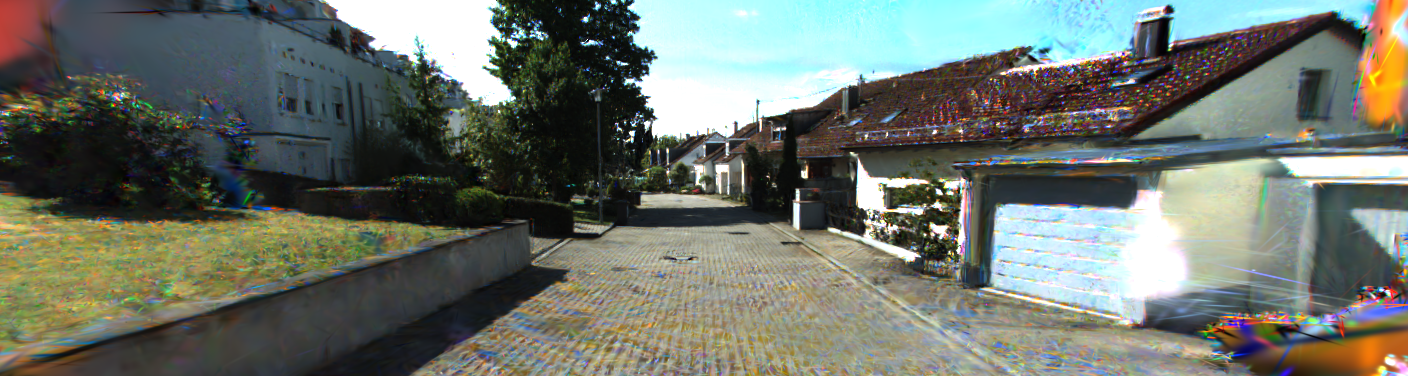}
      &
      \includegraphics[width=\mywidth\linewidth]{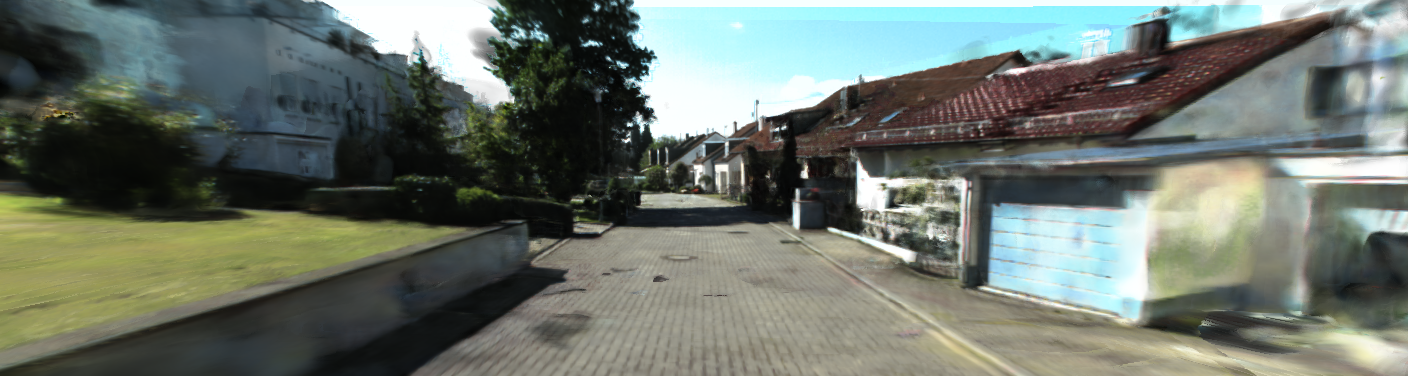}
      &
      \includegraphics[width=\mywidth\linewidth]{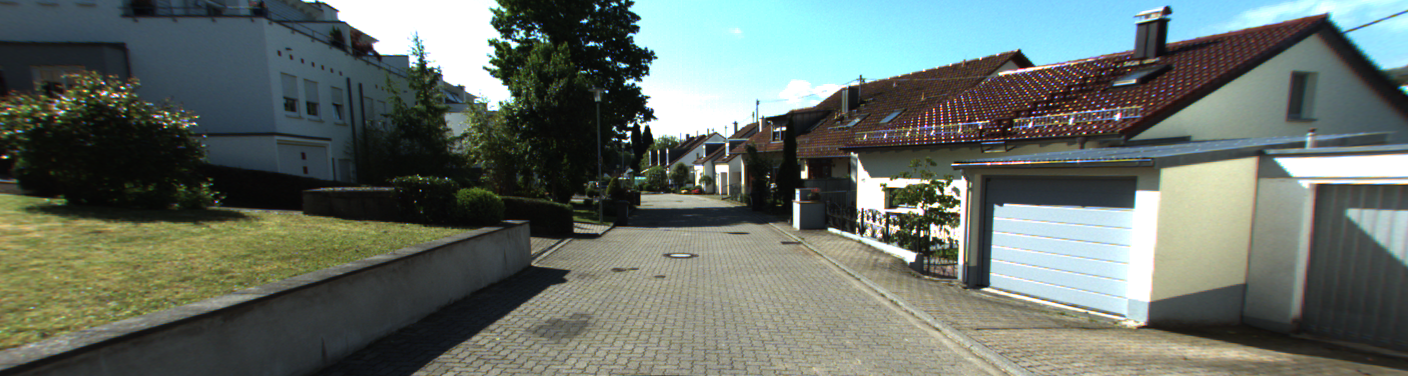}
      \\
      3DGS & Ours & GT
    \end{tabular}
 \caption{{\bf Qualitative Comparison} with Test-Time Optimization baselines.}  
 \label{fig: test time}
\end{figure}

We proceed to compare our method with test-time optimization approaches across various sparsity levels. \cref{tab:test time} and \cref{fig: test time} demonstrate that our method achieves state-of-the-art performance at high sparsity levels, attributable to the robust initialization provided by our generalizable model. In the drop 50\% setting, the recent state-of-the-art method 3DGS \cite{3dgs} achieves the best SSIM and LPIPS, but their performance experiences significant degradation in the 80\% and 90\% settings.
Our method achieves superior performance than other geometric regularization\cite{mixnerf} or depth-supervision\cite{sparsenerf,ds-nerf} methods. Although they show good performance on small-scale datasets (\eg DTU, LLFF datasets), they struggle with urban scenes which contain more complex geometry layouts. In a sparser setting, we maintain our advantages and achieve a PSNR of 20.91dB in 80\% drop rate and 19.16dB PSNR in 90\% drop rate, respectively. Another advantage of our work is that we achieve good quality with fast fine-tuning on novel scenes. Given the same computation recourse, our training time (\textbf{5 minutes}) outperforms other baselines by a large margin, $5\times$ faster than DS-Nerf\cite{ds-nerf} and SparseNerf\cite{sparsenerf} and $10\times$ faster than MixNeRF\cite{mixnerf}. 

\begin{table}[tb]  
\begin{tabular}{m{2.0cm}<{\centering} m{2.4cm}<{\centering}  m{1.8cm}<{\centering}  m{1.9cm}<{\centering} m{1.15cm} m{1.15cm} m{1.15cm}}
    \toprule
    \centering
  Setting & Methods & Drop Rate &Cost Time & PSNR$\uparrow$  & SSIM$\uparrow$ & LPIPS$\downarrow$  \\
    \midrule
    \multirow{9}{*}{\begin{minipage}{2cm} \centering  Geometric Reg.  \end{minipage}}  & \multirow{3}{*}{MixNeRF}  & 
    50\% & $\sim{51min}$ & 21.50 & 0.625 & 0.357 \\
    & & 80\% & $\sim{38min}$ & 18.89 & 0.558 & 0.404 \\
    & & 90\% & $\sim{31min}$ & 17.89 & 0.529 &  0.416   \\[0.05cm] \cline{3-7}
    
    \cline{3-7}
    & \multirow{3}{*}{SparseNeRF}    & 50\% & $\sim{35min}$ & 21.34 & 0.642 & 0.552 \\
    & & 80\% & $\sim{27min}$ & 19.18 & 0.607 & 0.559 \\
    & & 90\% & $\sim{21min}$ & 17.94 & 0.559 &  0.608 \\[0.05cm]  \cline{3-7}  
    
    \cline{3-7}
    & \multirow{3}{*}{DS-NeRF}  &  50\% & $\sim{36min}$ & 20.40 & 0.604 & 0.601 \\
    & & 80\% & $\sim{28min}$ & 18.91 & 0.581 & 0.620 \\
    & & 90\% & $\sim{22min}$ & 17.91 & 0.564 & 0.632 \\[0.05cm] %

    \midrule
    \multirow{6}{*}{\begin{minipage}{2cm} \centering  Depth Input  \end{minipage}}   
    & \multirow{3}{*}{3D GS}  &  50\% & $\sim{29min}$ & 24.37 & \textbf{0.804} & \textbf{0.109} \\
    & & 80\% & $\sim{18min}$ & 19.80 & 0.679 & 0.234 \\
    & & 90\% & $\sim{10min}$ & 17.46 & 0.597 &0.332 \\[0.05cm] \cline{3-7}

    \cline{3-7}
    & \multirow{3}{*}{\textbf{Ours}}  &  50\% & $\sim{5min}$ & \textbf{24.43} & 0.793 & 0.136 \\
    & &  80\% &  $< 5min$ &  \textbf{20.91} &  \textbf{0.712} & \textbf{0.220} \\
    & & 90\% &  $< 5min$ & \textbf{19.16} & \textbf{0.657} & \textbf{0.271} \\

    \bottomrule
    \end{tabular}
    \captionof{table}{\textbf{Test-Time Optimization Comparison.}}
    \label{tab:test time}
    \end{table}
    
\subsection{Ablation Study}
In \cref{tab:ablation}, we ablate the design choices of our method. We report the results on KITTI-360 validation dataset with 50\% drop rate. First, we replace the modulated SPADE CNN  with the traditional 3D U-Net, yielding the results (3D U-Net). This worsens feed-forward inference results, indicating that the modulated convolutions better preserve the information of the input point cloud. Next, we remove the 3D feature from the full model, yielding a baseline similar to IBRNet but augmented with scene decomposition (w/o $\bff_\text{fg}^\text{3D}$). This leads to a significant drop in performance, indicating the importance of the geometric priors. We also study the effect of the 2D image-based feature, and observe that removing it leads to blurry rendering (w/o $\bff_\text{fg}^\text{2D}$). This demonstrates that the 3D feature $\bff_\text{fg}^\text{3D}$ alone is insufficient due to the inherent inductive bias of the 3D CNN. When removing the street decomposition and representing the entire scene as foreground (w/o decomposition), it leads to artifacts in background regions due to the incorrect geometric modeling. Finally, we demonstrate that the LiDAR supervision of the training sequences leads to slight improvements (w/o $\cL_{lidar}$). We observe that the LiDAR supervision improves details, leading to a larger improvement in LPIPS compared to PSNR and SSIM. Note that LiDAR supervision is not applied during test, but only applied to the training sequences to enhance our generalizable geometry prediction against noisy geometric priors. Please refer to our supplementary material for qualitative comparisons.

\begin{table}[t]
\label{tab:table2}
    \centering
    \begin{tabular}{m{4cm}<{\centering}  m{1.05cm} m{1.05cm} m{1.05cm} <{\centering} m{0.3cm} m{1.05cm} m{1.05cm} m{1.05cm} <{\centering}}
    \toprule
    \multirow{2}{*}{Ablation} & \multicolumn{3}{c}{Feed-foward NVS} & & \multicolumn{3}{c}{Finetuning NVS}\\
    \cline{2-4}  \cline{6-8}  
    &PSNR$\uparrow$ & SSIM$\uparrow$ & LPIPS$\downarrow$ & & PSNR$\uparrow$ & SSIM$\uparrow$ & LPIPS$\downarrow$ \\

    \midrule
    3D U-Net  &21.49   &0.716   &0.194  & &24.14  &0.789  &0.145 \\
    w/o $\bff_\text{fg}^\text{3D}$ &15.71   &0.532   &0.441  & &18.91  &0.609  &0.343 \\
    w/o $\bff_\text{fg}^\text{2D}$  & 18.28  &0.593   &0.342  & &21.41  &0.648  &0.315 \\
    w/o decomposition &20.65   & 0.684    &0.195  & & 22.54 & 0.754 &0.175 \\
    w/o $\cL_{lidar}$ & 21.75   &0.742   &0.189 & &23.77 &0.788  &0.139   \\
    \midrule
    Full model  &21.93    &0.745   &0.178   &  &  24.43&0.793  &0.136  \\
    
    \bottomrule
    \end{tabular}
    \caption{\textbf{Ablation study.} Metrics are averaged over the 5 test scenes.}
    \label{tab:ablation}
\end{table}

\section{Conclusion}
The paper presents \method, a generalizable and efficient method for sparse urban view synthesis. By integrating geometric priors into a generalizable model, we demonstrate that our method achieves robust performance across various density levels and datasets. It further enables efficient fine-tuning and outperforms existing sparse view methods based on good initialization. We will investigate handling dynamic objects in street views in future work, which poses challenges to generalizable approaches.

\section*{Acknowledgements}
 We would like to thank Dzmitry Tsishkou for his useful suggestions and tips. This work was supported by the National Key R\&D Program of China under Grant 2021ZD0114500, and NSFC under grant 62202418 and U21B2004. A. Geiger was supported by the ERC Starting Grant LEGO-3D (850533) and the DFG EXC number 2064/1 - project number 390727645.

\bibliographystyle{splncs04}
\bibliography{egbib}
\newpage
\begin{center}
\Large{\textbf{Supplementary Material for \\Efficient Depth-Guided Urban View Synthesis}}
\end{center}
\appendix
\begin{abstract}

    In this {\bf supplementary document}, we initially present an in-depth overview of our network architecture in \cref{sec: network arch}. Subsequently, we elaborate on our implementation details in \cref{sec: imp details}, encompassing the random masking strategy, the sampling strategy, and the point cloud accumulation configurations. Following this, we introduce the baselines for comparison in \cref{sec: baselines}, and discuss the datasets utilized in our experiments in \cref{sec: datasets}. Finally, we present additional experimental results in \cref{sec: additional exp}. Besides, the {\bf supplementary video} showcases our feed-forward and fine-tuned results across various scenes, along with novel view renderings compared against multiple baselines.

\end{abstract}

\renewcommand\thesection{\Alph{section}}
\section{Network Architecture}
\label{sec: network arch}
\subsection{Modulation-based CNN}
\begin{figure*}[h]
    \centering
    \def\mywidth{.33}
    \includegraphics[width=\linewidth]{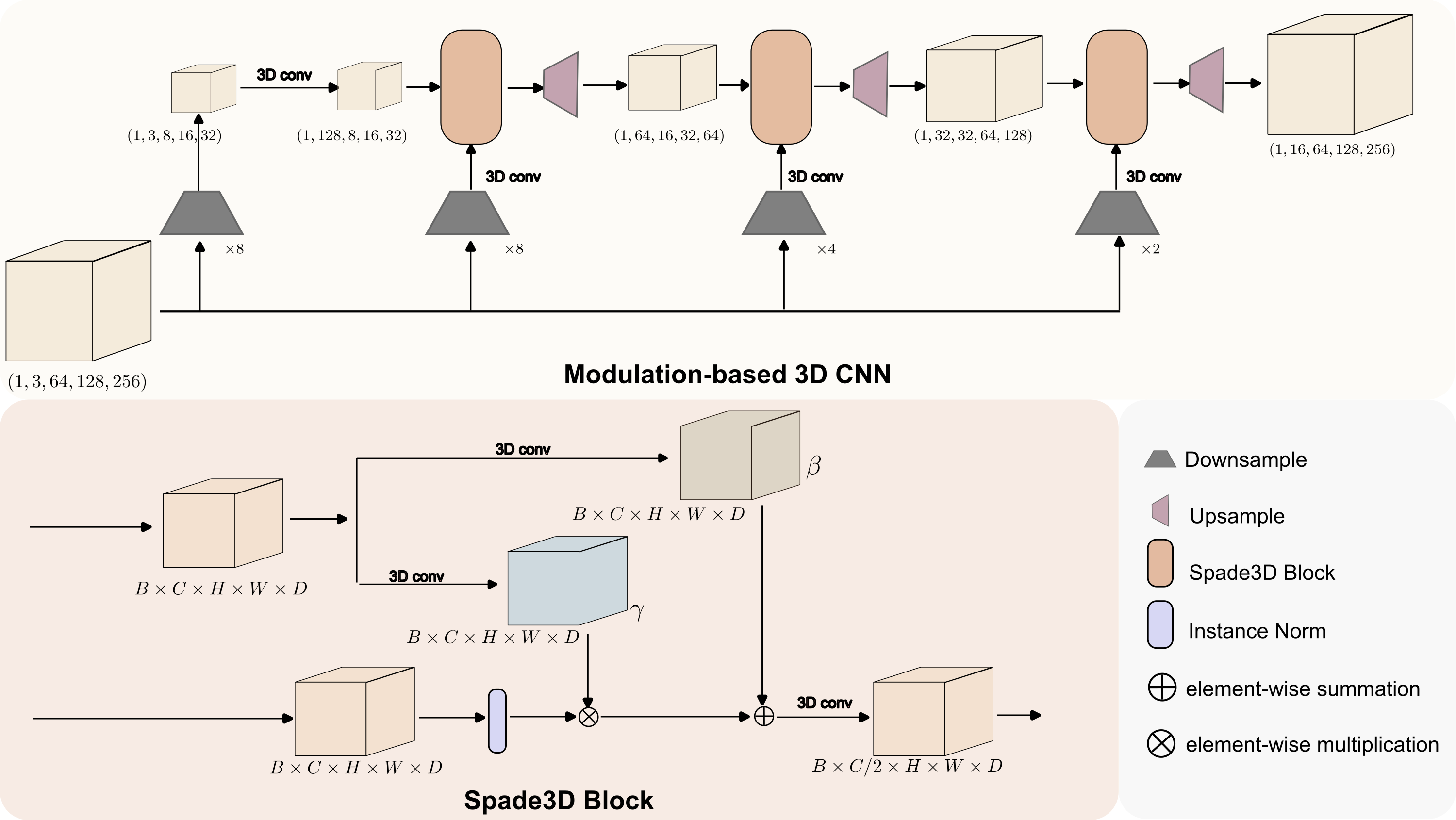}
    \caption{\textbf{Architecture} of our 3D SPADE CNN.}
    \label{fig:spade}
\end{figure*}

Our 3D Spatially-Adaptive Normalization convolutional neural network (SPADE CNN) modulates the feature volume generation at multiple resolutions. It takes a single voxelized point cloud with dimension $64\times128\times256$ as input. The appearance information of the input point cloud is injected into the feature volume at each resolution.  \cref{fig:spade} presents the detailed network architecture.

\subsection{Foreground Network Architecture}
The feature volume generated by the 3D SPADE CNN provides both geometric priors and appearance information for foreground contents. Based on a feature vector $\bff_\text{fg}^\text{3D}\in\nR^{16}$ extracted from this feature volume, our density decoder $g_\theta$ predicts the density value. In order to capture high-frequency appearance during feedforward inference, the color decoder $h_\theta$ combines the 3D feature $\bff_\text{fg}^\text{3D}\in\nR^{16}$ and 2D features queried from nearest views $\bff_\text{fg}^\text{2D}\in\nR^{9}$ for color prediction.
In addition, the color decoder further takes the positional encoding of $\bx$ as input, setting the number of frequencies to 10 following \cite{nerf}.
Furthermore, we inject viewing direction $\bd\in\nR^{3}$ and a per-frame appearance embedding $\bw\in\nR^{32}$.
We present the detailed network architecture in \cref{tab:Network architecture}.

\begin{table}[h!]
  \centering
 
    \begin{tabular}{m{2.20cm}  m{2.20cm}<{\centering} m{2.20cm}<{\centering} m{3.0cm}<{\centering}}
    \toprule
    \multicolumn{4}{l}{$\textbf{Density Decoder}$} \\
    \hline
    \textbf{Layer} &\textbf{in channels} &\textbf{out channels} &\textbf{description} \\
    LinearRelu$_{0}$ &  16 & 64 & in: $\bff_\text{fg}^\text{3D}$ \\
    LinearRelu$_{1}$ & 64 & 1 & out: $\sigma_\text{fg}$  \\
    \hline
    \multicolumn{4}{l}{$\textbf{Color Decoder}$} \\
    \hline
    \textbf{Layer} &\textbf{in channels} &\textbf{out channels} &\textbf{description} \\
    LinearRelu$_{0}$ & 16+9+63 & 128 & in: $\bff_\text{fg}^\text{3D}, \bff_\text{fg}^\text{2D}, \gamma(\bx)$ \\
    LinearRelu$_{i=1,2}$ & 128 & 128 & / \\
    LinearRelu$_{3}$ & 128+32+3 & 128 & in: + $\textbf{d}, \bw$ \\
    LinearRelu$_{4}$ & 128 & 64 & / \\
    LinearRelu$_{5}$ & 64 & 3 & out: $\bc_\text{fg}$ \\
     \bottomrule
    \end{tabular}%
    \vspace{0.1cm}
    \caption{\textbf{Density and Color Decoder Architectures.} From top to bottom: two-layer MLP for density prediction and six-layer MLP for color prediction.} 
  \label{tab:Network architecture}%
\end{table}%

\subsection{Background and Sky Network Architectures}
We also build the generalizable model for the background and sky region via the image-based rendering paradigm as we mentioned in the main paper. We now show the details of the background and sky network details in \cref{tab: background network architecture}.

\begin{table}[h]
  \centering
 
    \begin{tabular}{m{2.20cm}  m{2.20cm}<{\centering} m{2.20cm}<{\centering} m{3.0cm}<{\centering}}
    \toprule
    \multicolumn{4}{l}{$\textbf{Background network architecture}$} \\
    \hline
    \textbf{Layer} &\textbf{in channels} &\textbf{out channels} &\textbf{description} \\
    LinearRelu$_{0}$ &  63+9 & 128 & in: $\bff_\text{fg}^\text{2D}, \gamma(\bx)$ \\
    LinearRelu$_{i=1,2,3}$ & 128 & 128 & /  \\
    LinearRelu$_{4}$ & 128 & 16 & out: $\sigma_\text{bg}$, $\bf{f}_{e}$ \\
    LinearRelu$_{5}$ & 15+3 & 64 & / \\
    LinearRelu$_{6}$ & 64 & 64 & / \\
    LinearRelu$_{7}$ & 64 & 3 & out: $\bc_\text{bg}$ \\
    \hline
    \multicolumn{4}{l}{$\textbf{Sky network architecture}$} \\
    \hline
    \textbf{Layer} &\textbf{in channels} &\textbf{out channels} &\textbf{description} \\
    LinearRelu$_{0}$ & 9+3 & 3 & in: $\bff_\text{fg}^\text{2D}, \textbf{d}$; out: $\bc_\text{sky}$\\
    
     \bottomrule
    \end{tabular}%
    \vspace{0.1cm}
    \caption{\textbf{Background and Sky Network Architectures}. The $\bf{f}_{e}\in\nR^{15}$ represents the geometric embedding vector and serves as the input for background color prediction.} 
  \label{tab: background network architecture}%
\end{table}%

\section{Implementation Details}
\label{sec: imp details}
\subsection{Input Volume Random Masking}
Our target is to perform novel view synthesis from sparse urban images.
When provided with sparse images, the accumulated point clouds are often incomplete and contain holes due to occlusions or insufficient overlap across the input frames. To equip EDUS with the capability to handle sparse and incomplete point clouds by filling in the missing parts, we randomly mask portions of the point cloud during the training stage. Specifically, we randomly select an $8\mathrm{m}\times8\mathrm{m}\times12\mathrm{m}$ cuboid in each iteration, corresponding to a volume of size $[40\times40\times60]$, and eliminate the point cloud within this area. Note that this is not applied in feed-forward inference nor per-scene fine-tuning.
This simple yet effective strategy enhances the generalizable ability on novel scenes, as shown in our ablation study in \cref{sec:ablation_masking}.

\subsection{Hierarchical Sampling}
The total number of samples located at the ray is 160 with 128 sampled from the foreground and 32 from the background.
Specifically, we first select 80 points along the ray within the foreground volume, distributing one-half of these samples uniformly, while the other half is allocated linearly based on disparity spacing, following Nerfacto~\cite{tancik2023nerfstudio}. Next, we iteratively perform importance sampling to sample valid regions containing solid contents based on the coarse density prediction following NeuS~\cite{wang2021neus}, with three iterations and 16 samples for each iteration. For the background, we additionally uniformly sample 32 points out of the foreground volume.

\subsection{Accumulated Point Cloud}
We design our accumulated point cloud to encourage the model to learn completion. Considering that the sparsity level in real-world driving data is unknown in advance, we aim to train a \textit{single} generalizable model and evaluate its performance at different sparsity levels. In our main experiments, all generalizable methods are trained at the sparsity level of the drop50 setting, i.e., 50\% images are available for supervision. Instead of directly using the 50\% reference images to create the accumulated point cloud, we create a more sparse point cloud for training to encourage the model to learn scene completion. Specifically, the accumulated point cloud we use for training is from every fifth of the stereo pairs. Note that during the feed-forward inference or per-scene optimization, we always use the given reference images of different sparsity levels (50\%, 80\%, or 90\%) to generate the corresponding accumulated point cloud. This ensures fair comparison to the baselines and further verifies our method's robustness to the density of the input point cloud.

\section{Baselines}
\label{sec: baselines}
\subsubsection{Generalizable NeRFs}
We utilize the official implementations of IBR-Net, MVSNeRF, Neo360, and MuRF. For each method, we pretrain the model using 80 scenes from the KITTI360 dataset \cite{liao2022kitti} under a 50\% drop rate. In the case of IBR-Net, MVSNeRF, and MuRF, we select the nearest three training frames of the target view as reference frames. As MVSNeRF constructs local cost volumes defined at reference views, we adjust the fine-tuning strategy of MVSNeRF to dynamically use the nearest reference frames to construct the cost volume, similar to the pre-training phase, rather than employing a fixed set of frames as in their original implementation. 

\subsubsection{Test-Time Optimization Methods}
The test-time optimization approaches are individually trained for each scene under three sparsity levels. We employ the official codebase of MixNeRF \cite{mixnerf} and implement SparseNeRF \cite{sparsenerf} and DS-NeRF \cite{ds-nerf} by integrating stereo depth supervision onto Nerfstudio's Mip-NeRF model \cite{tancik2023nerfstudio, mipnerf}. Additionally, we utilize the splatfacto model of Nerfstudio to train our 3D GS \cite{3dgs}.
To ensure a fair comparison, we provide the depth maps used in our approach to train the depth-supervision methods \cite{sparsenerf, ds-nerf}. When it comes to point cloud-based methods \cite{3dgs}, we find that the missing of sky may lead to a decline of performance, so we keep the sky in the point cloud and initialize the model with it.

\section{Datasets}
\label{sec: datasets}
\subsection{KITTI-360} 
We follow the 50\% drop rate (Drop50) and 90\% drop rate (Drop90) settings of the KITTI-360 dataset and expand them into 80\% drop rate (Drop80). To be more specific, we take a pair of left and right eye image pairs as the unit. In the Drop50 setting, we use odd-numbered image pairs as the training set. Similarly, in the Drop80 and Drop90 settings, we use every fifth or every tenth pair of images for training, respectively. Note that all test frames are the same to allow comparison across different sparsity levels. For example, considering ten consecutive image pairs, Drop50 refers to using pairs of id $0,2,4,6,8$ for training, whereas Drop80 and Drop90 refer to using $0,5$ and $0$, respectively. The test frames are selected as the left-eye image of $1,3,7,9$ pairs to avoid overlapping with the training frames.

\subsection{Waymo}
We use the same drop rate setting and point cloud accumulation scheme as the KITTI-360 dataset for Waymo. Note that we only use the front camera of the Waymo dataset. Therefore, every odd-numbered image is used for training for the Drop50 setting, and every fifth or tenth image is used for Drop80 and Drop90, respectively. The test views are configured to encompass frames at the 3rd and 7th positions, as well as frames with id equivalent to 3 or 7 plus the multiples of 10, which has no overlapping with the training set.

\section{Additional Experimental Results}
\label{sec: additional exp}
\subsection{Qualitative Results of Ablation Study}
\begin{figure}[t]
    \centering
    \setlength{\tabcolsep}{1pt}
    \def\mywidth{1.0}
    \def\mywidthP{6.0cm}
    \begin{tabular}{cc}
        \begin{subfigure}{0.45\textwidth}
            \centering
            \includegraphics[width=\linewidth]{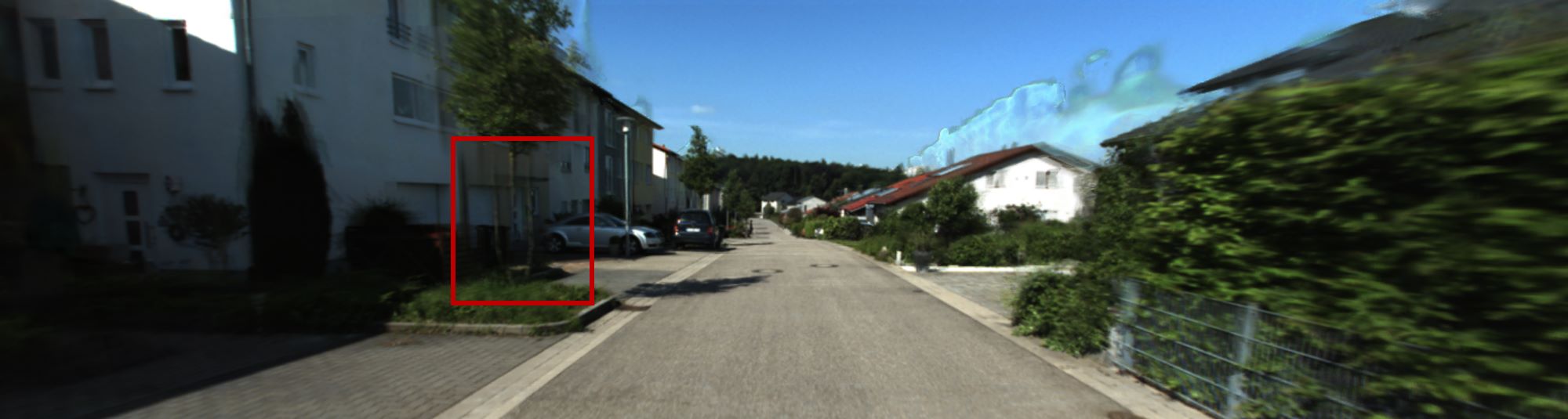}
            \caption{3D U-Net}
        \end{subfigure}
        &
        \begin{subfigure}{0.45\textwidth}
            \centering
            \includegraphics[width=\linewidth]{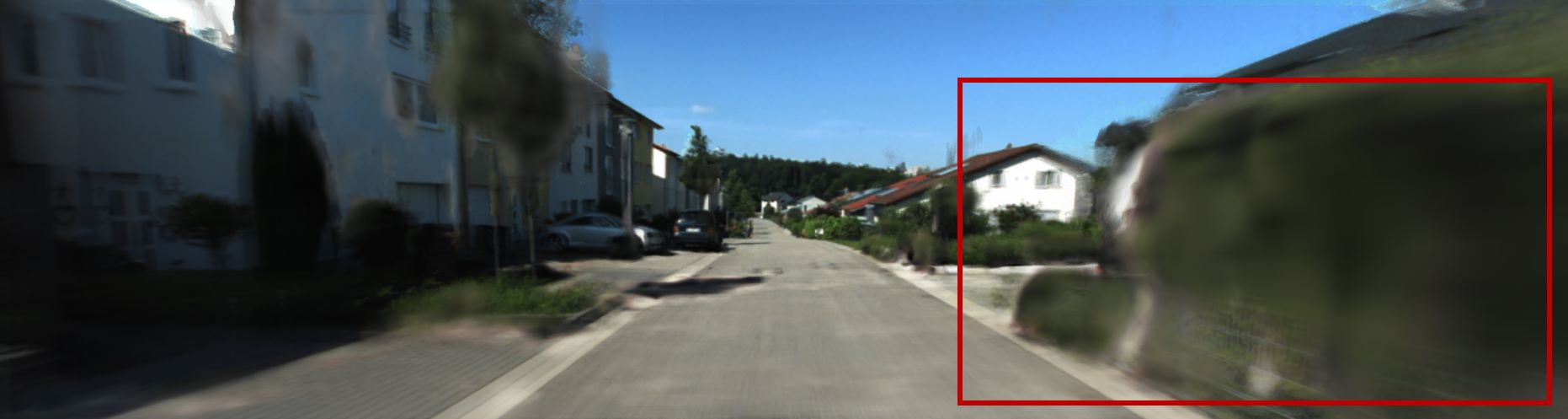}
            \vspace{-0.44cm}
            \caption{w/o $\bff_\text{fg}^\text{2D}$}
        \end{subfigure}
        \\
        \begin{subfigure}{0.45\textwidth}
            \centering
            \includegraphics[width=\linewidth]{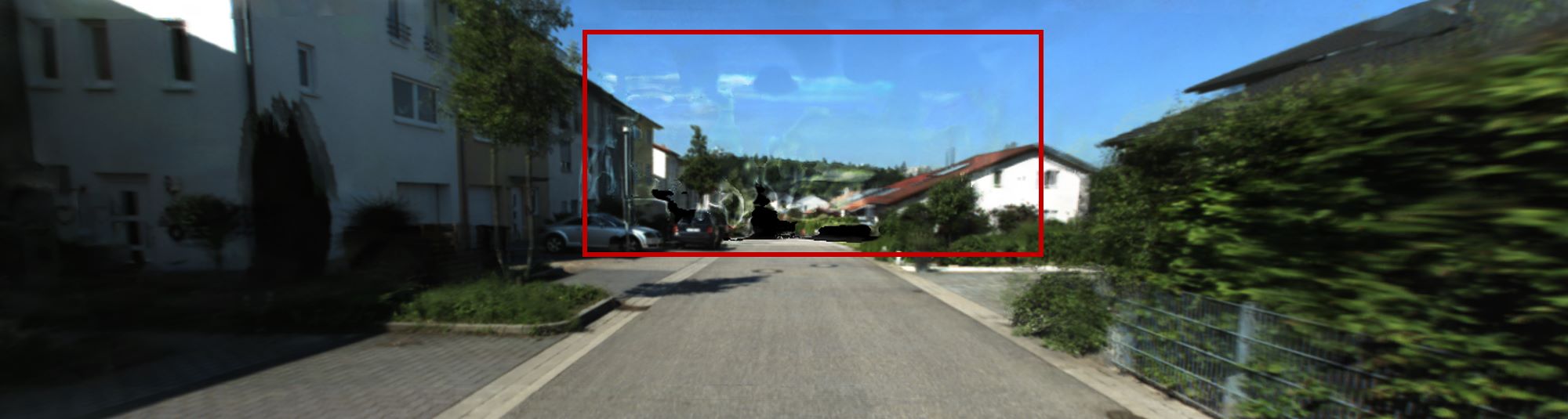}
            \caption{w/o decomposition}
        \end{subfigure}
        &
        \begin{subfigure}{0.45\textwidth}
            \centering
            \includegraphics[width=\linewidth]{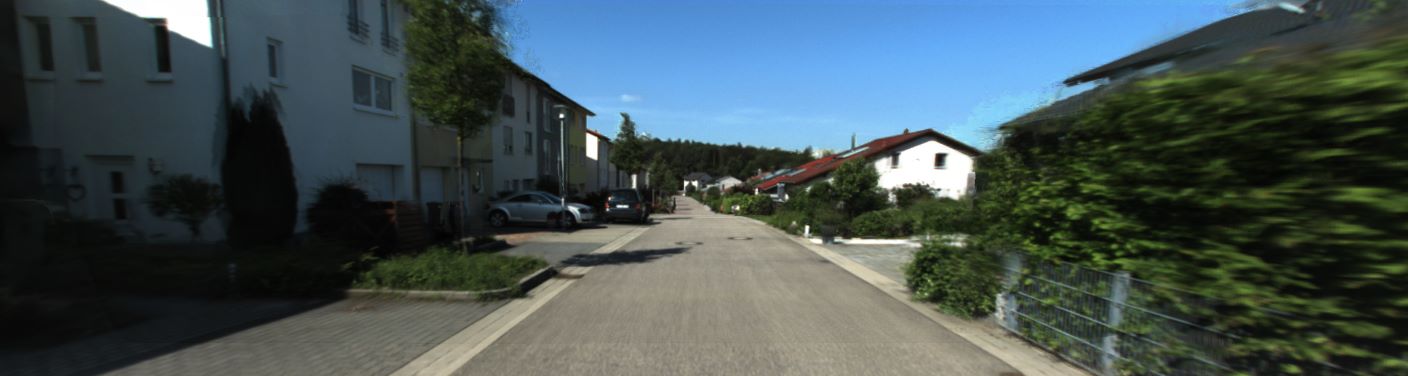}
            \caption{Full model}
        \end{subfigure}
        \\
        \multicolumn{2}{c}{\begin{subfigure}{0.6\textwidth}
            \centering
            \includegraphics[width=0.8\linewidth]{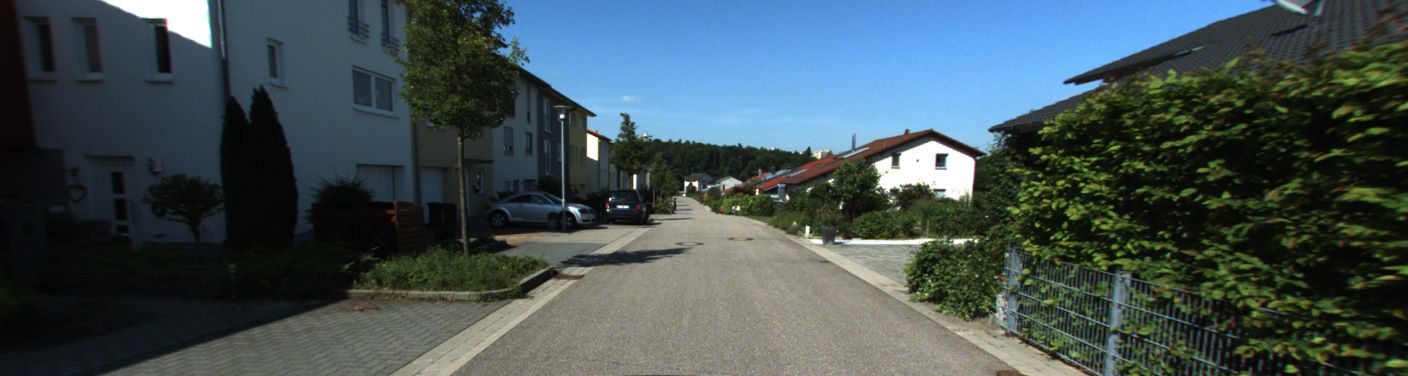}
            \caption{GT}
        \end{subfigure}}
    \end{tabular}
    \caption{{\bf Qualitative Results of Ablation Study.} The experiments are conducted on KITTI-360 under Drop50 sparsity level.}  
    \label{fig:supp_ablation}
\end{figure}

\cref{fig:supp_ablation} demonstrates the qualitative outcomes of our ablation study. These results are all derived from feed-forward inference images under the Drop50 sparsity level, which align seamlessly with the corresponding quantitative results in Table 3 of the main paper. The substitution of the modulated SPADE CNN with a conventional 3D U-Net leads to less detailed prediction and yields blurriness in some regions. The elimination of the 2D image-based feature results in the high-frequency contents losing their sharpness and becoming noticeably blurry.
Moreover, if we solely represent the scene as foreground, it results in artifacts in the distant region, due to the wrong modeling of the scene geometry. Our fully integrated model exhibits the highest quality of reconstruction, effectively validating our design choices.

\subsection{Additional Ablation of Input Volume Random Masking}
\label{sec:ablation_masking}
As shown in \cref{tab:ablation}, the random mask strategy helps improve our feed-forward performance, which means our model is equipped with filling capacity to process the incomplete point cloud by adding random masking. 

\begin{table}[t]
    \centering
    \begin{tabular}{m{4cm}<{\centering}  m{1.05cm} m{1.05cm} m{1.05cm} <{\centering} m{0.3cm} m{1.05cm} m{1.05cm} m{1.05cm} <{\centering}}
    \toprule
    
    \multirow{2}{*}{Ablation} & \multicolumn{3}{c}{KITTI$360_{drop50}$} & & \multicolumn{3}{c}{KITTI$360_{drop80}$}\\
    \cline{2-4}  \cline{6-8} &
    PSNR$\uparrow$ & SSIM$\uparrow$ & LPIPS$\downarrow$ & & PSNR$\uparrow$ & SSIM$\uparrow$ & LPIPS$\downarrow$ \\

    \midrule
    w/o random masking &20.62   &0.694   &0.228  & &18.54  &0.641  &0.288 \\
   
    \midrule
    Full model  & 21.93 & 0.745 & 0.178 &  & 19.63 & 0.668 & 0.244  \\
    
    \bottomrule
    \end{tabular}
    \vspace{0.2cm}
    \caption{\textbf{Ablation Study of Input Volume Random Masking.} The results show that the random masking strategy in the training stage improves the feed-forward inference performance using different sparse settings.}
    \label{tab:ablation}
\end{table}

\subsection{Additional Comparison on Waymo Dataset}
\begin{figure}[t]
    \centering
    \begin{tabular}{ccc}
        \begin{subfigure}{0.3\textwidth}
            \centering
            \includegraphics[width=\linewidth]{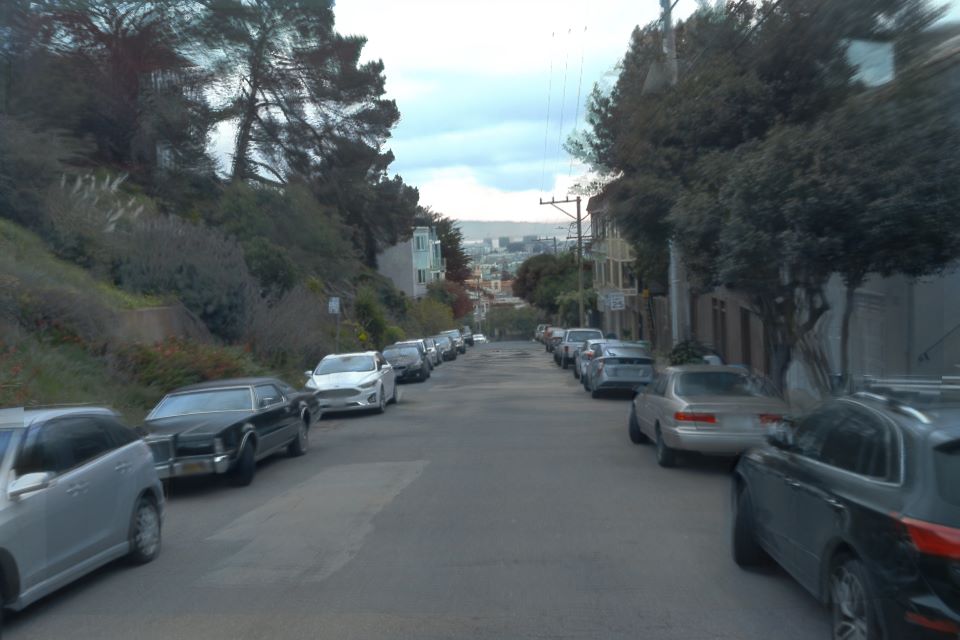}
        \end{subfigure}
        &
        \begin{subfigure}{0.3\textwidth}
            \centering
            \includegraphics[width=\linewidth]{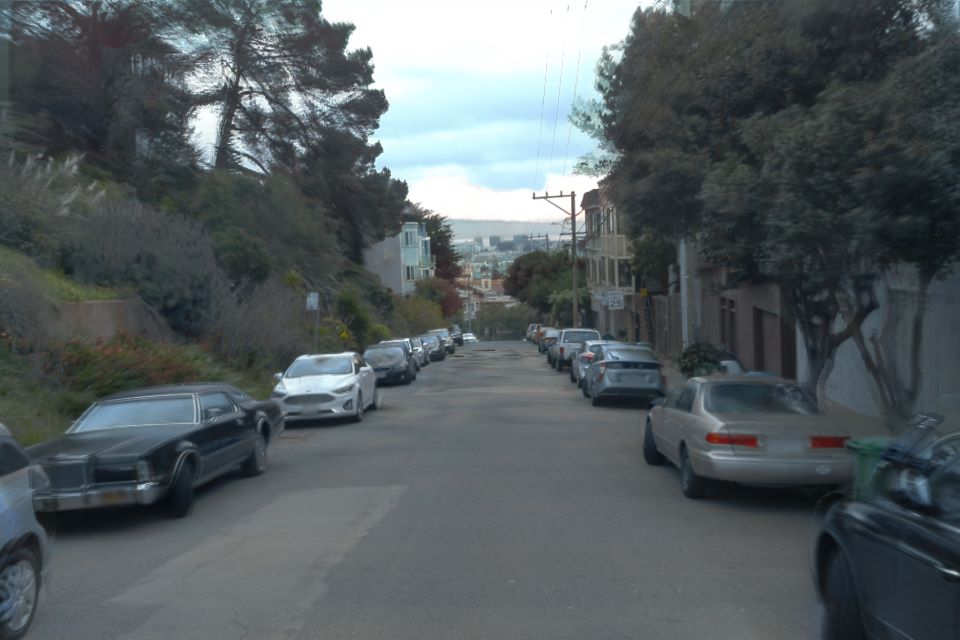}
        \end{subfigure}
        &
        \begin{subfigure}{0.3\textwidth}
            \centering
            \includegraphics[width=\linewidth]{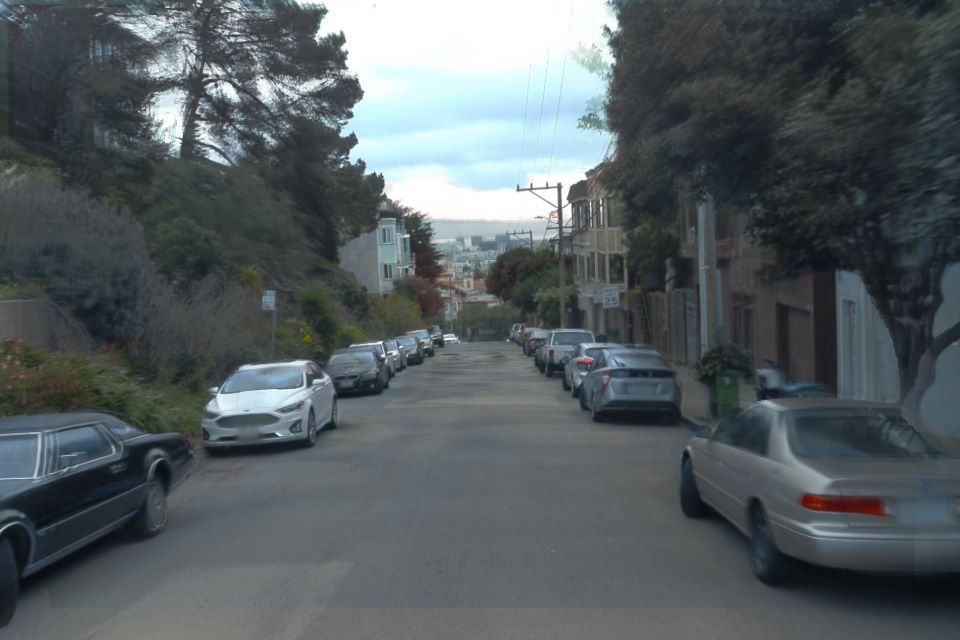}
        \end{subfigure}
        \\
        \begin{subfigure}{0.3\textwidth}
            \centering
            \includegraphics[width=\linewidth]{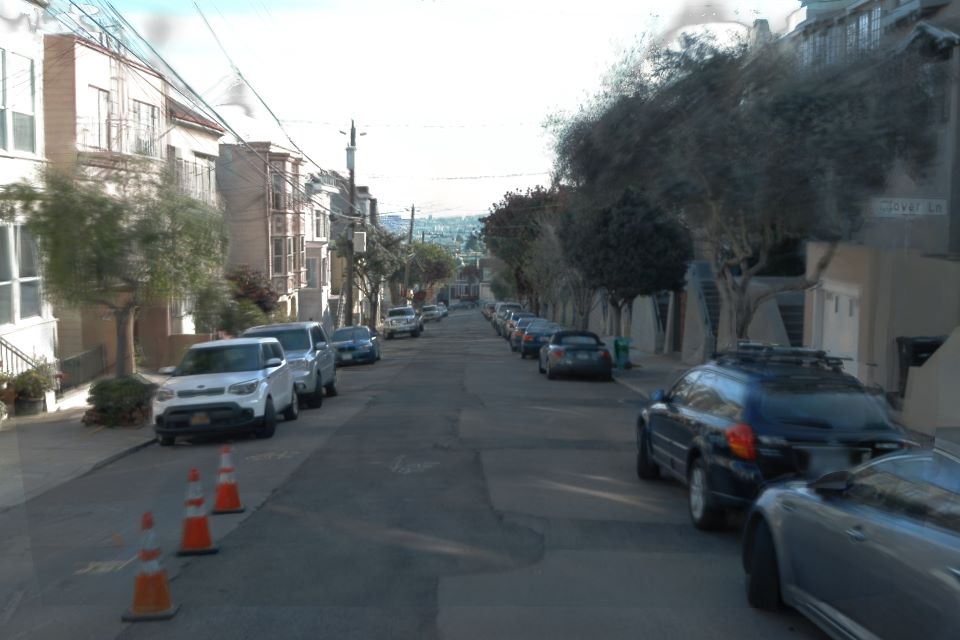}
        \end{subfigure}
        &
        \begin{subfigure}{0.3\textwidth}
            \centering
            \includegraphics[width=\linewidth]{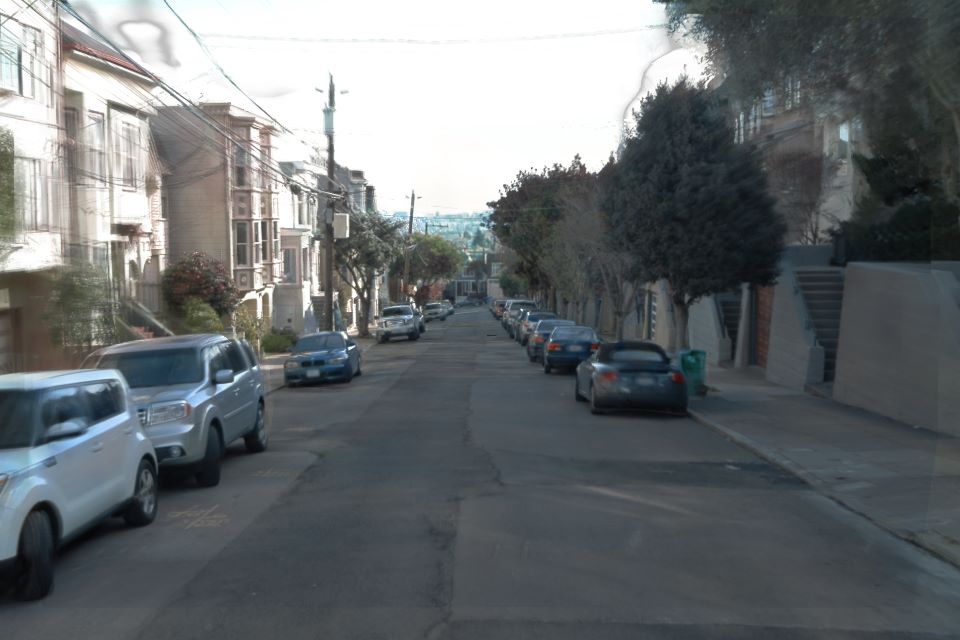}
        \end{subfigure}
        &
        \begin{subfigure}{0.3\textwidth}
            \centering
            \includegraphics[width=\linewidth]{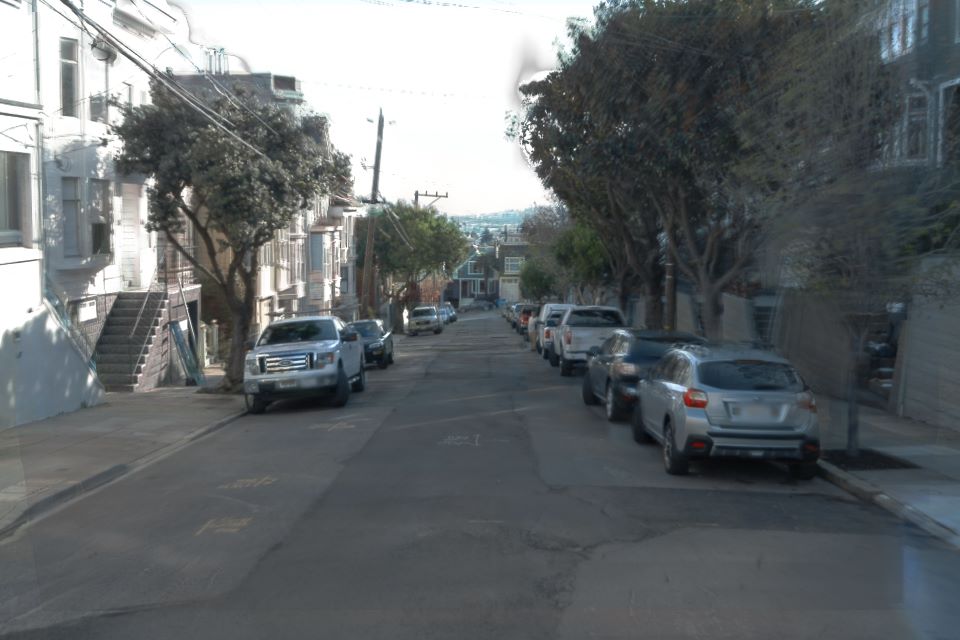}
        \end{subfigure}
    \end{tabular}
    \caption{{\bf Zero-shot Inference} on Waymo dataset. Our model is trained on the KITTI-360 dataset using Drop50 sparsity level.}
    \label{fig:waymo zero-shot}
\end{figure}

\begin{figure}[h!]
    \centering
    \begin{tabular}{P{0.4cm}P{5cm}P{5cm}}
        \rotatebox{90}{Feed-forward} &
        \includegraphics[width=0.4\textwidth]{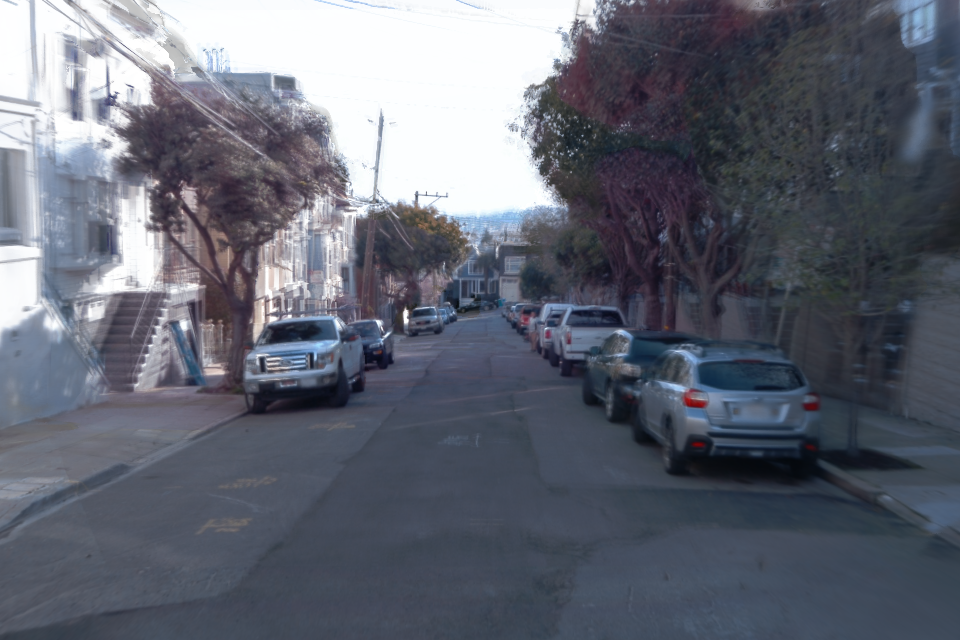} &
        \includegraphics[width=0.4\textwidth]{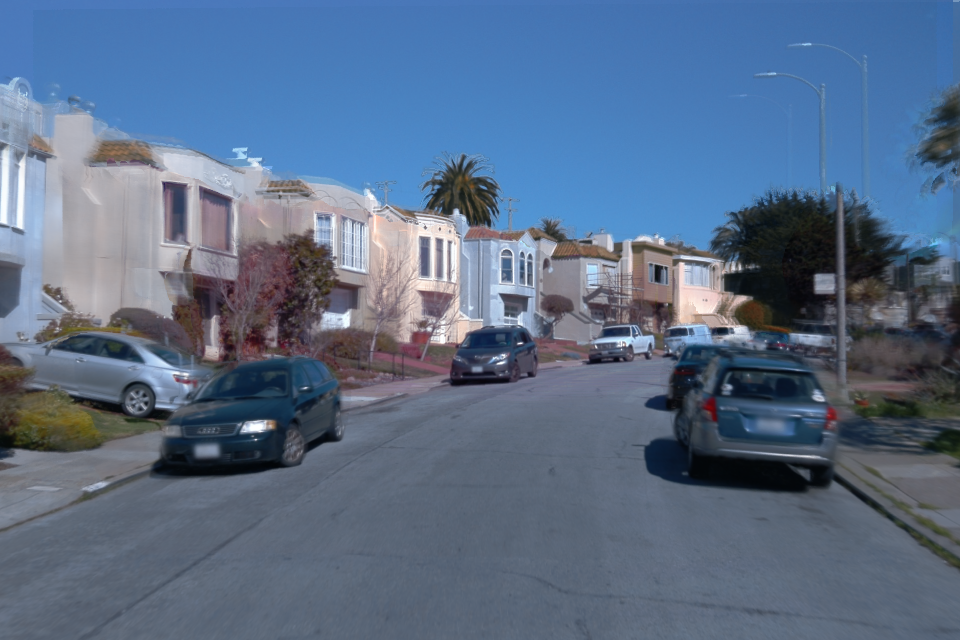} \\
        \rotatebox{90}{Fine-tuned} &
        \includegraphics[width=0.4\textwidth]{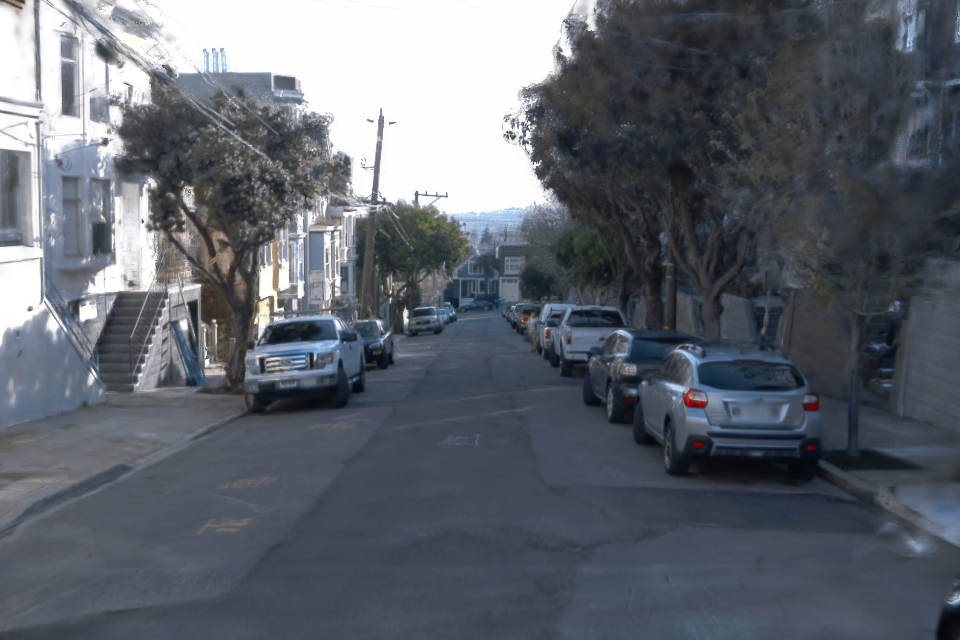} &
        \includegraphics[width=0.4\textwidth]{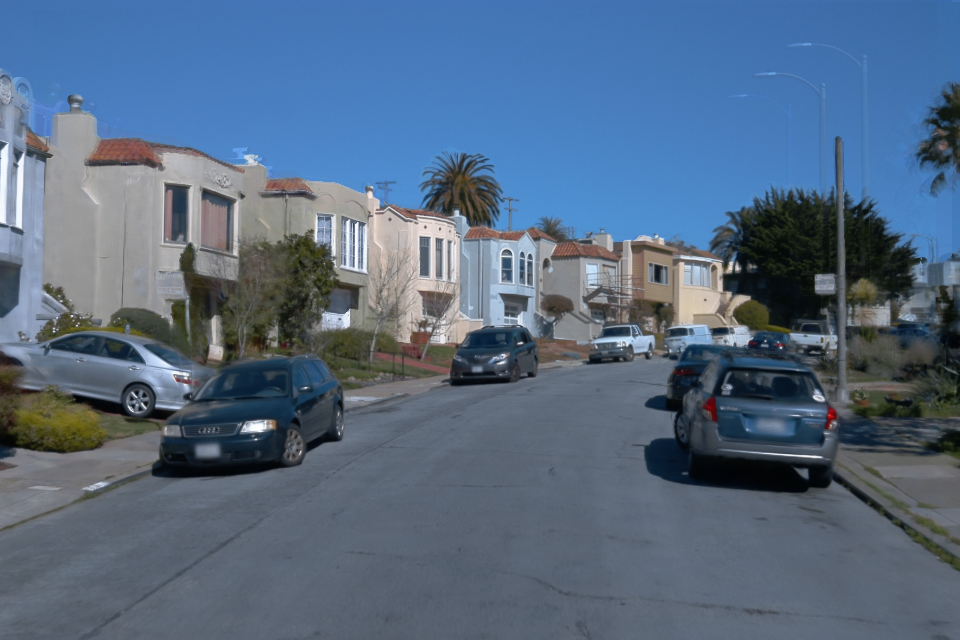}\\
    \end{tabular}
    \caption{{\bf Qualitative Results} trained and evaluated on Waymo dataset. Our model is trained on Waymo dataset using Drop50 sparsity level. The feed-forward inference and fine-tuning are performed under the Drop80 setting.}
    \label{fig: waymo_drop80}
\end{figure}

We conduct zero-shot NVS under a drop rate of 50\% using the model trained on the KITTI-360 dataset, and the rendered images are shown in \cref{fig:waymo zero-shot}. To further verify our method's performance given different training scenes, we train all generalizable methods on the Waymo dataset and evaluate them on Waymo and KITTI-360, see \cref{tab:waymo_generalizable}. Specifically, all models are trained with the Drop50 setting on the Waymo dataset. Our model has better performance under 80\% drop rate thanks to the depth-guided global volume and achieves the best metrics after fine-tuning.  Note that our depth guidance of the Waymo dataset is obtained from a monocular depth estimation method~\cite{yin2023metric3d}, demonstrating that we are able to learn an effective generalizable model guided by noisy monocular depth priors. \cref{fig: waymo_drop80} shows the feed-forward and fine-tuned qualitative results of our method trained and evaluated on the Waymo dataset.

\begin{table}[t]
  \centering
   \resizebox{!}{2.4cm}{
  \begin{tabular}{m{2.0cm} m{1.5cm}<{\centering} m{0.95cm}<{\centering} m{0.95cm}<{\centering} m{0.95cm}<{\centering} m{0.15cm} m{0.95cm}<{\centering} m{0.95cm}<{\centering} m{0.95cm}<{\centering} m{0.15cm} m{0.95cm}<{\centering} m{0.95cm}<{\centering} m{0.95cm}<{\centering}}
    \toprule
    
    \multirow{2}{*}{Methods} & \multirow{2}{*}{Setting} & \multicolumn{3}{c}{Waymo$_{drop50}$} & & \multicolumn{3}{c}{Waymo$_{drop80}$} & & \multicolumn{3}{c}{KITTI360$_{drop50}$} \\ \cline{3-5}  \cline{7-9}  \cline{11-13}
  & & PSNR$\uparrow$  & SSIM$\uparrow$  & LPIPS$\downarrow$ 
  & & PSNR$\uparrow$  & SSIM$\uparrow$  & LPIPS$\downarrow$ 
  & & PSNR$\uparrow$  & SSIM$\uparrow$  & LPIPS$\downarrow$ \\
    \midrule
    IBR-Net  & \multirow{5}{*}{\begin{minipage}{1.5cm} \centering No \\ per-scene \\ opt. \end{minipage}}  & 22.90 & 0.739 & 0.148 & & 17.90 & 0.619 & 0.242 & & 18.88 & 0.606 & 0.272 \\
    MVSNeRF &  & 18.05 & 0.595 & 0.378 & & 17.41 & 0.581 & 0.404 & & 15.35 & 0.524 & 0.389 \\
    Neo360 &  & 16.28 & 0.505 & 0.569 & & 15.83 & 0.505 & 0.584 & & 10.99 & 0.204 & 0.687 \\
    MuRF &  & \textbf{23.66} & 0.746 & 0.256 & & 21.25 & 0.664 & 0.327 & & 19.83 & \textbf{0.669} & 0.340  \\
    \textbf{Ours} &  & 23.41 & \textbf{0.769} & \textbf{0.147} & & \textbf{21.64} & \textbf{0.703} & \textbf{0.204} & & \textbf{20.13} & 0.659  & \textbf{0.257}  \\[0.1cm] 
    \cline{1-13} 
     IBR-Net & \multirow{5}{*}{\begin{minipage}{1.5cm} \centering Per-scene \\ opt. \end{minipage}} & 23.71 & \textbf{0.861} & 0.121 & & 18.58 & 0.741 & 0.222 & & 20.78 & 0.650 & 0.205 \\
    MVSNeRF &  & 22.42 & 0.699 & 0.309 & & 20.81 & 0.657 & 0.348 & & 18.80 & 0.611 & 0.354 \\
    Neo360 &  & 21.76 & 0.647 & 0.517 & & 20.66 & 0.631 & 0.533 & & 17.89 & 0.480 & 0.571 \\
    MuRF &  & 24.06 & 0.753 & 0.224 & & 22.85 & 0.711 & 0.260 & & 21.24 & 0.710 & 0.265 \\ 
    \textbf{Ours} &  & \textbf{26.88} & 0.839 & \textbf{0.109} & & \textbf{24.33} & \textbf{0.776} & \textbf{0.164} & & \textbf{24.35} & \textbf{0.787} & \textbf{0.145} \\
  \bottomrule
  \end{tabular}
  }
  \vspace{0.2cm}
  \caption{\textbf{Quantitative Comparison} on five test scenes among generalizable methods. All models are trained on the Waymo dataset using Drop50 sparsity level.
  }
\label{tab:waymo_generalizable}
\vspace{0.2cm}
\end{table}

\subsection{Additional Qualitative Analysis of Geometry Prediction}
\begin{figure}[h!]
    \centering
    \begin{tabular}{cc}
        \begin{subfigure}{0.45\textwidth}
            \centering
            \includegraphics[width=\linewidth]{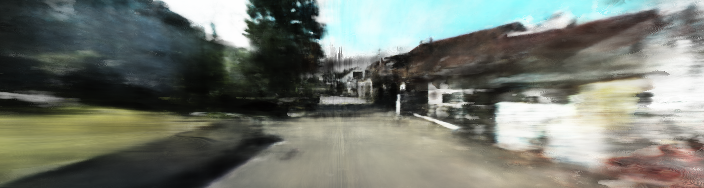}
        \end{subfigure}
        &
        \begin{subfigure}{0.45\textwidth}
            \centering
            \includegraphics[width=\linewidth]{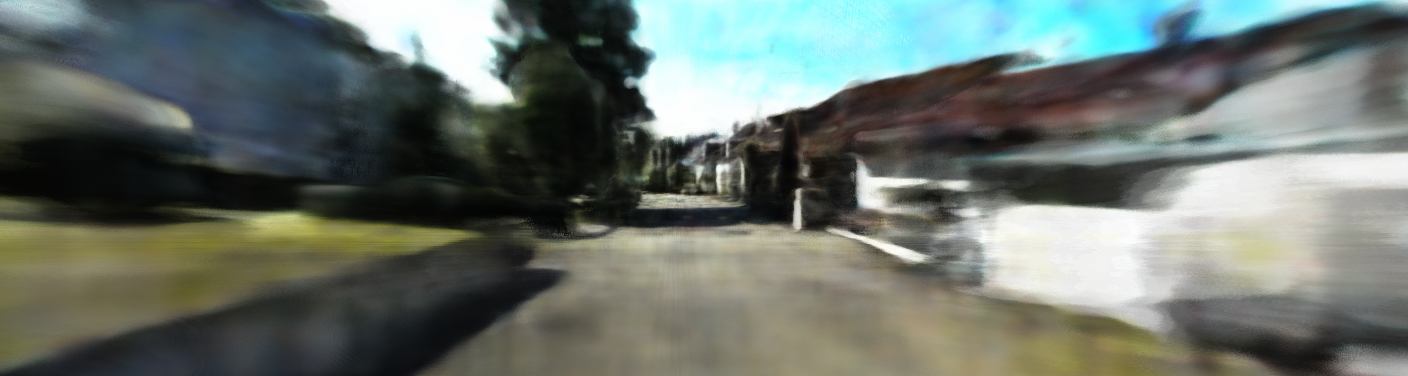}
        \end{subfigure}
        \\
        \begin{subfigure}{0.45\textwidth}
            \centering
            \includegraphics[width=\linewidth]{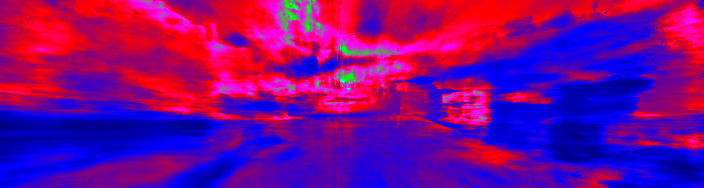}
            \caption{Mix-NeRF}
        \end{subfigure}
        &
        \begin{subfigure}{0.45\textwidth}
            \centering
            \includegraphics[width=\linewidth]{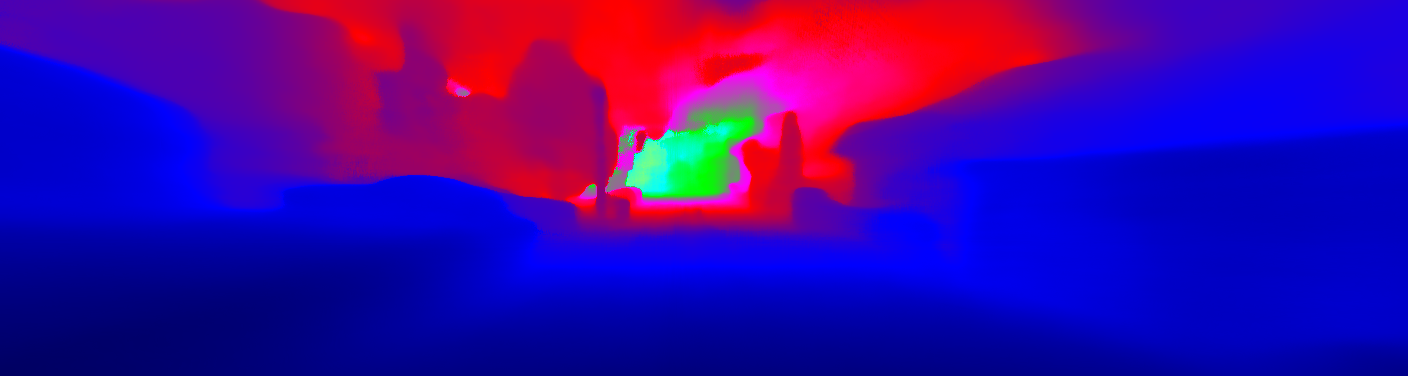}
            \caption{DS-NeRF}
        \end{subfigure}
        \\
        \begin{subfigure}{0.45\textwidth}
            \centering
            \includegraphics[width=\linewidth]{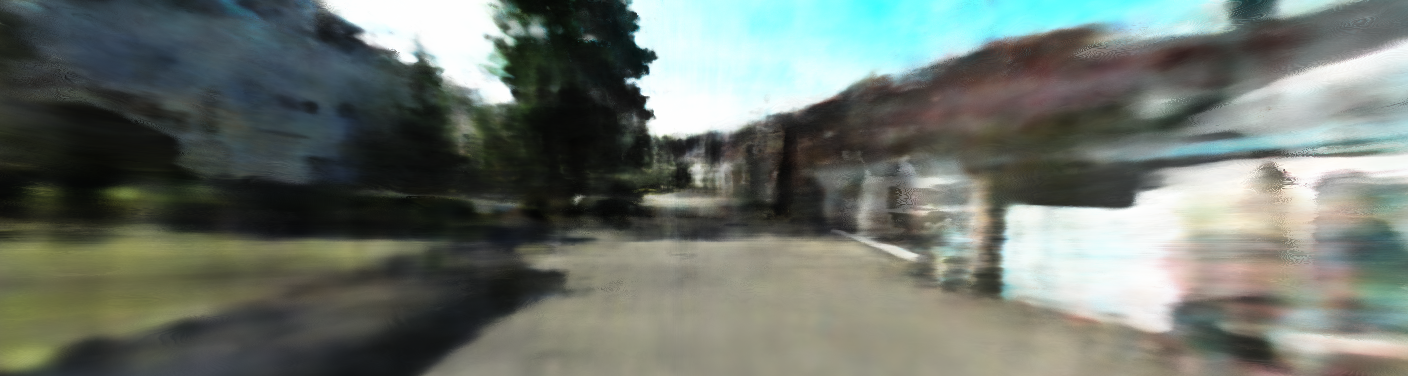}
        \end{subfigure}
        &
        \begin{subfigure}{0.45\textwidth}
            \centering
            \includegraphics[width=\linewidth]{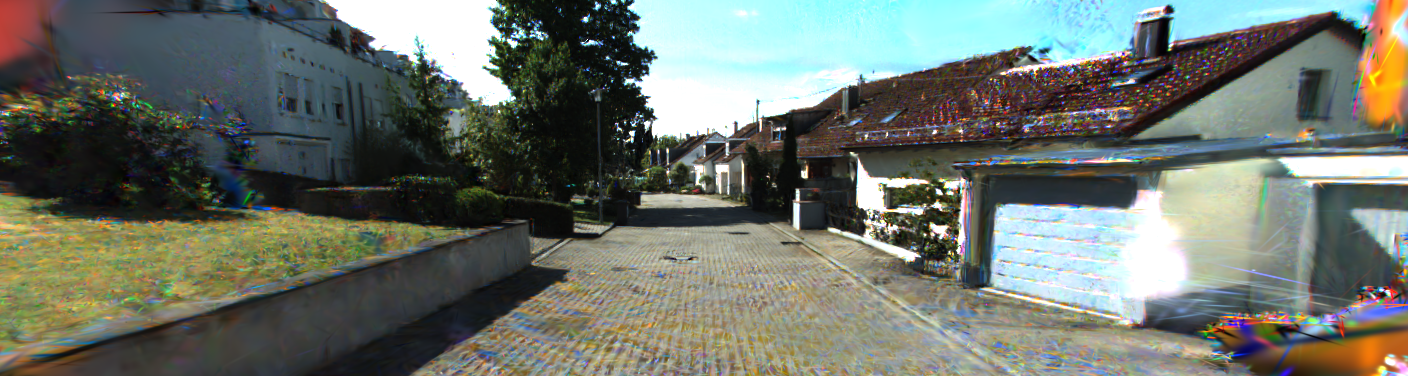}
        \end{subfigure}
        \\
        \begin{subfigure}{0.45\textwidth}
            \centering
            \includegraphics[width=\linewidth]{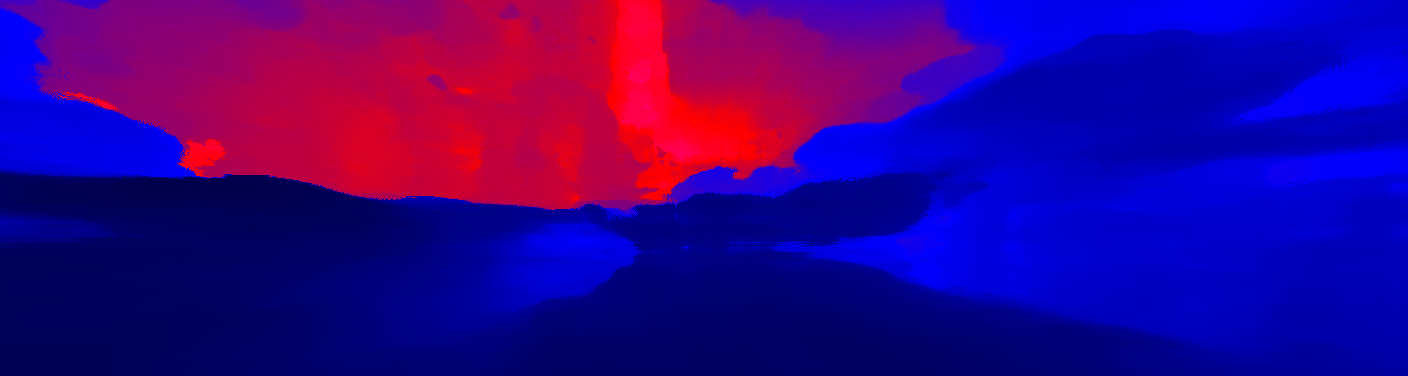}
            \caption{SparseNeRF}
        \end{subfigure}
        &
        \begin{subfigure}{0.45\textwidth}
            \centering
            \includegraphics[width=\linewidth]{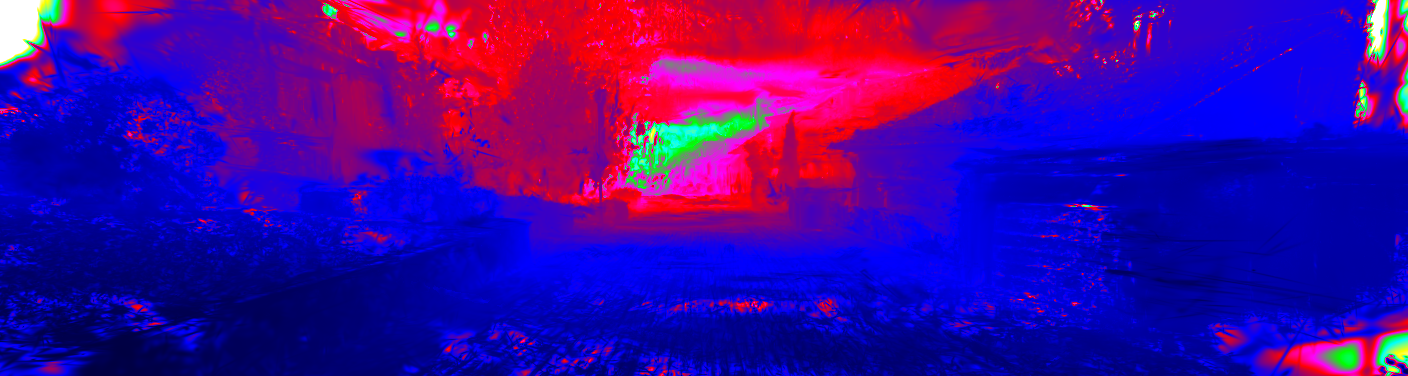}
            \caption{3DGS}
        \end{subfigure}
        \\
        \begin{subfigure}{0.45\textwidth}
            \centering
            \includegraphics[width=\linewidth]{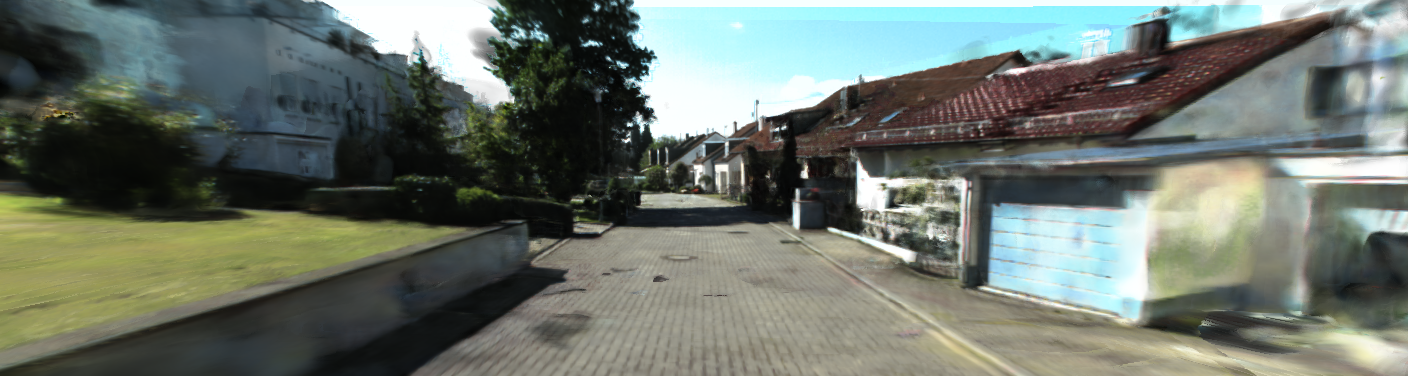}
        \end{subfigure}
        &
        \begin{subfigure}{0.45\textwidth}
            \centering
            \includegraphics[width=\linewidth]{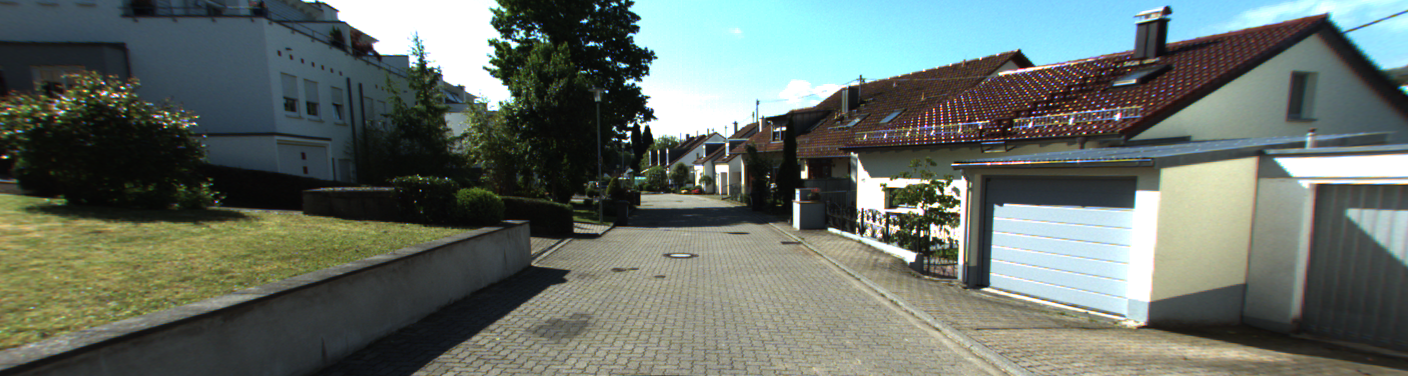}
        \end{subfigure}
        \\
        \begin{subfigure}{0.45\textwidth}
            \centering
            \includegraphics[width=\linewidth]{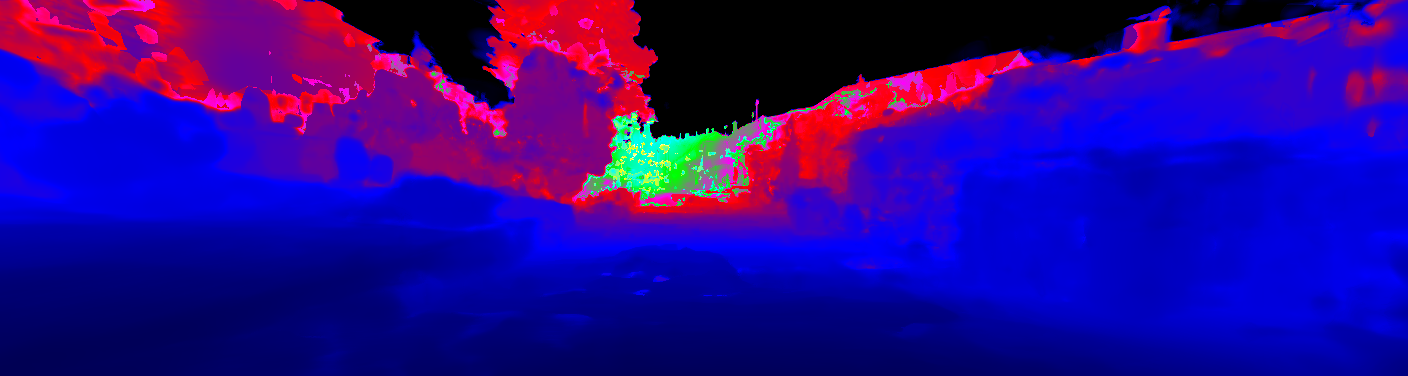}
            \caption{Ours}
        \end{subfigure}
        &
        \begin{subfigure}{0.45\textwidth}
            \centering
            \includegraphics[width=\linewidth]{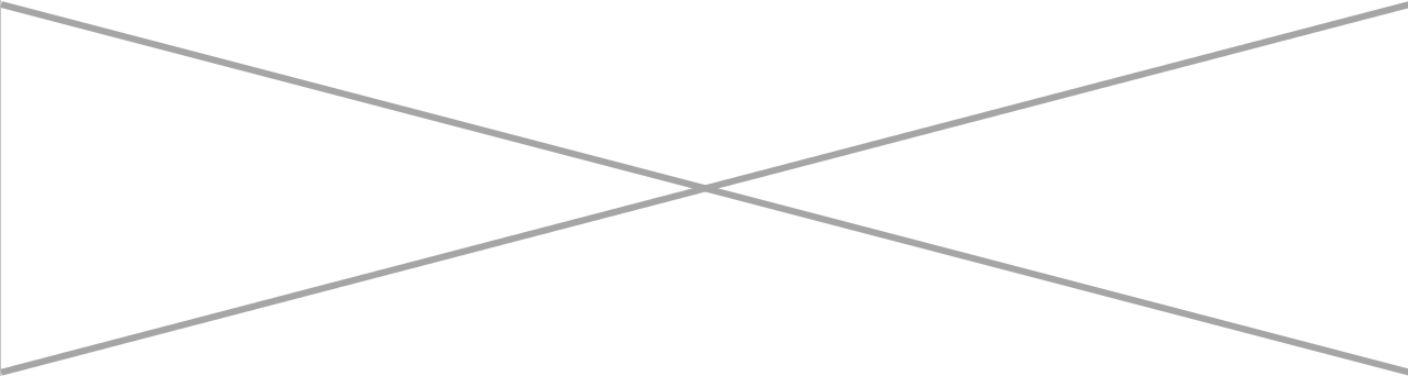}
            \caption{GT}
        \end{subfigure}
    \end{tabular}
    \caption{\textbf{Qualitative Depth Comparison} with test-time optimization methods.}
    \label{fig: test-time opt.}
\end{figure}

\begin{figure}[tbh]
    \centering
    \begin{tabular}{c}
        \begin{subfigure}{\textwidth}
            \centering
            \includegraphics[width=\linewidth]{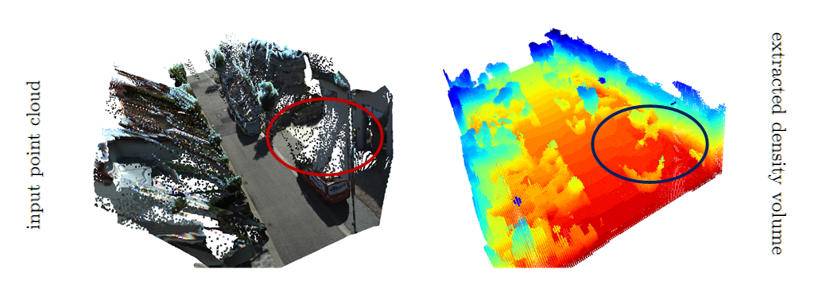}
            \caption{Filling holes in input point clouds.}
        \end{subfigure}
        \\
        \begin{subfigure}{\textwidth}
            \centering
            \includegraphics[width=\linewidth]{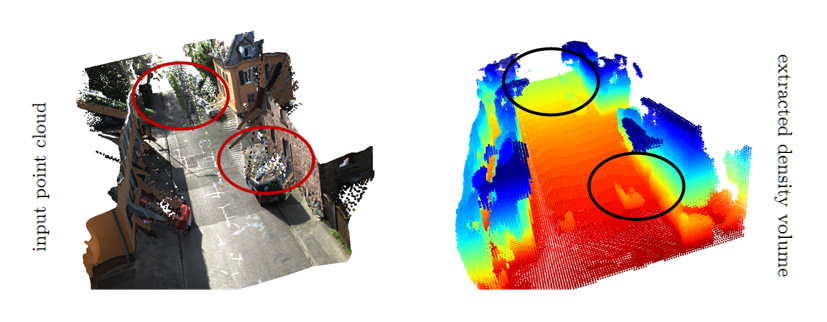}
            \caption{Removing noise from input point clouds.}
        \end{subfigure}
        \\
    \end{tabular}
    \caption{\textbf{Qualitative Results} of density volume extracted from our feed-forward inference.}
    \label{fig: feed-forward geometry}
\end{figure}

\cref{fig: test-time opt.} shows the novel view synthesis and depth comparison of our method to other optimization-based methods under Drop90 setting. The quantitative comparisons are reported in Table 2 of the main paper. 
Note that all methods except for Mix-NeRF use the same depth maps as supervision (DS-NeRF, SparseNeRF) or as input (3DGS, Ours). Although 3DGS\cite{3dgs} renders relatively good depth maps, it shows obvious flaws in appearance under sparse settings. Our model shows the best geometry thanks to our generalizable prior. As shown in \cref{fig: feed-forward geometry}, we illustrate the geometry extracted from the feed-forward inference, demonstrating that our EDUS is able to refine the noisy input point cloud and fill holes with generalizable 3D CNN.

\begin{table}[t]
    \centering
    \begin{tabular}{m{3.2cm}<{\centering} m{1.95cm}<{\centering} m{1.95cm}<{\centering} m{1.95cm}<{\centering}}
    \toprule
    
    Num. of Views & PSNR$\uparrow$ & SSIM$\uparrow$ & LPIPS$\downarrow$ \\
    \midrule
    
    2 & 18.06 & 0.606 & 0.374 \\
    3 & 18.69 & 0.639 & 0.353 \\
    4 & 18.47 & 0.633 & 0.361 \\
    6 & 16.61 & 0.589 & 0.410 \\
    8 & 15.01 & 0.544 & 0.459 \\
    10 & /    & /     & /     \\
    
    \bottomrule
    \end{tabular}
    \vspace{0.2cm}
    \caption{{\bf Number of Reference Views for MuRF.} We evaluate MuRF using various numbers of reference views. Note that the performance of MuRF does not increase wrt. the number of reference views.}
    \label{tab: num of reference views}
\end{table}

\subsection{Number of Reference Views for Generalizable NeRFs}
As we have mentioned in the main paper, our method takes input as a point cloud, which accumulates multi-frame information in addition to the reference frames. To verify fairness and our advancement compared with other reference-based methods, we evaluate the state-of-the-art approach MuRF\cite{xu2023murf} using different numbers of reference images. Specifically, we gradually increase the number of reference frames which is used for feature extraction and conduct feed-forward inference on our test sequence of the KITTI-360 dataset under Drop80 setting. We find that increasing the number of reference frames does not improve the model's performance. This may be due to the fact that further reference frames have stronger occlusions, hence increasing the difficulty of feature matching. On the other hand, the usage of more input frames also increases the burden on computing resources and causes out-of-memory when the number of views achieves 10. On the contrary, our method utilizes a fixed size of point cloud and manages to combine multi-frame guidance to predict accurate scene geometry. The results of MuRF using different numbers of reference views are shown in \cref{tab: num of reference views}.

\subsection{Limitations}
\begin{figure}[tb]
    \centering
    \begin{tabular}{P{5.5cm}P{5.5cm}}
        \includegraphics[width=0.45\textwidth]{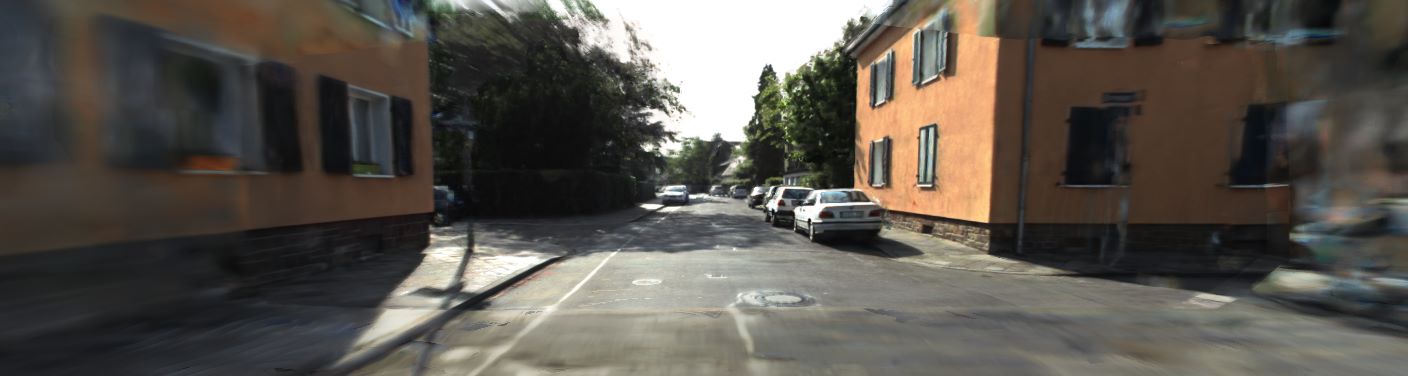} &
        \includegraphics[width=0.45\textwidth]{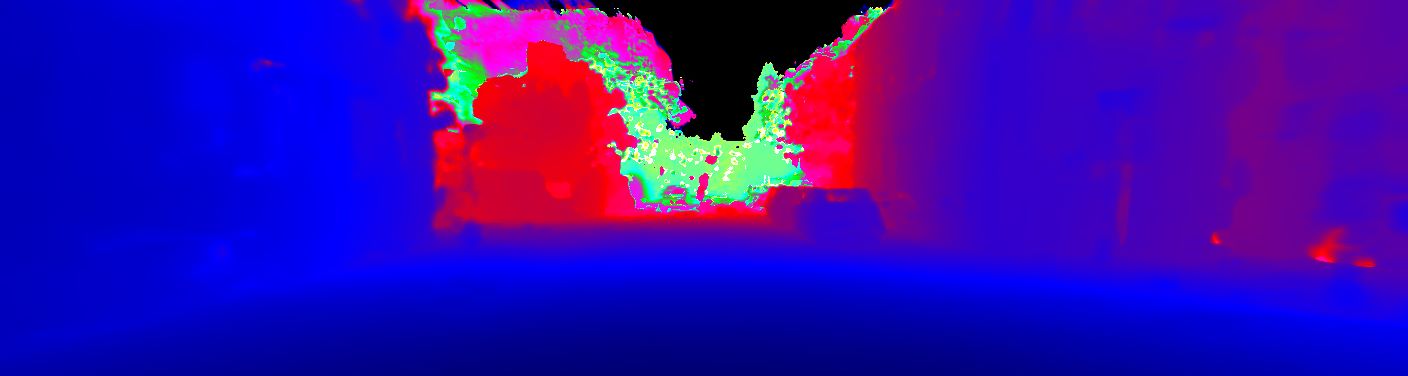} \\
        \includegraphics[width=0.45\textwidth]{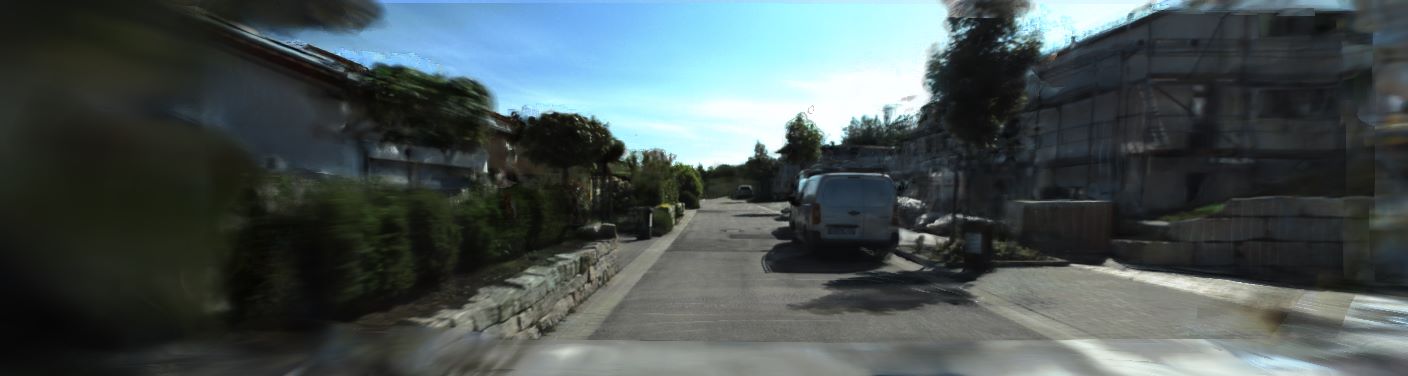} &
        \includegraphics[width=0.45\textwidth]{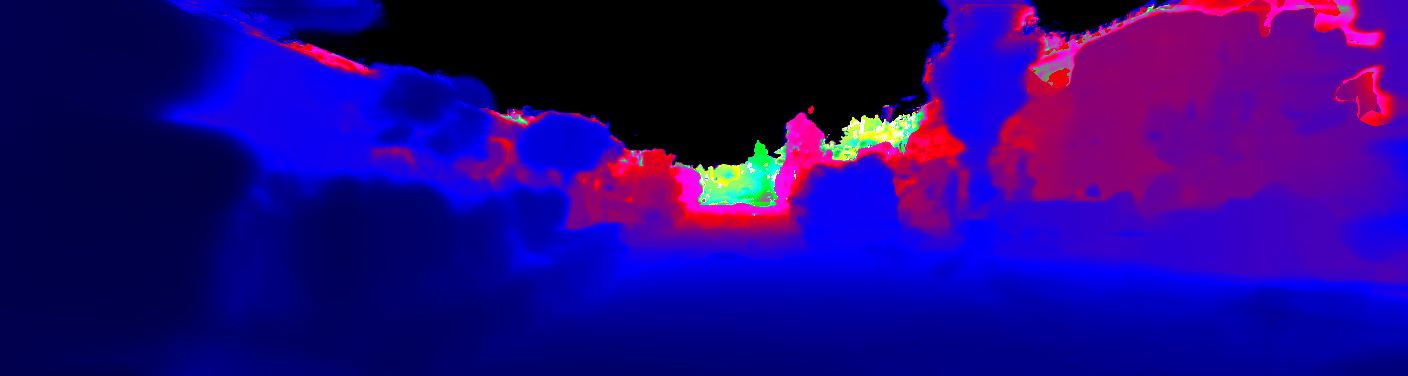} \\
        \includegraphics[width=0.45\textwidth]{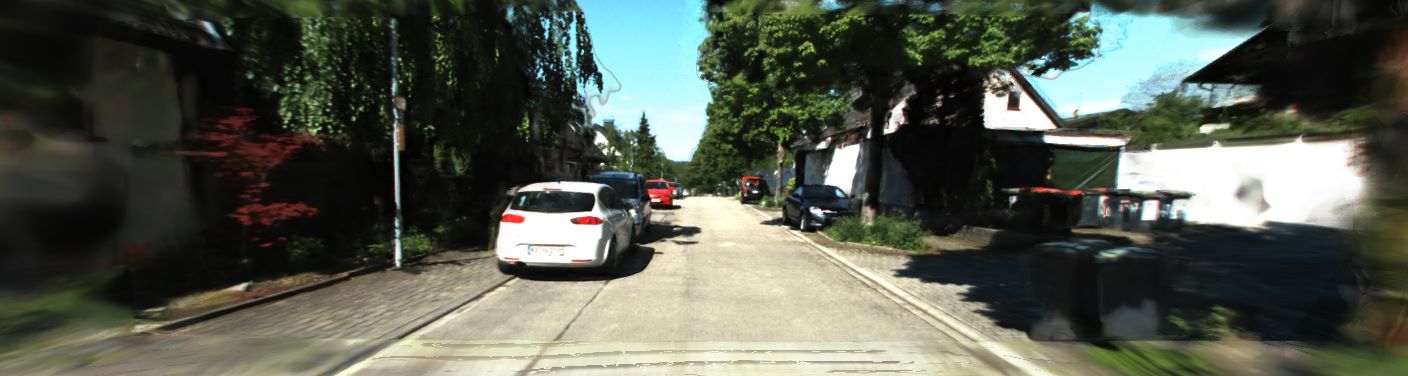} &
        \includegraphics[width=0.45\textwidth]{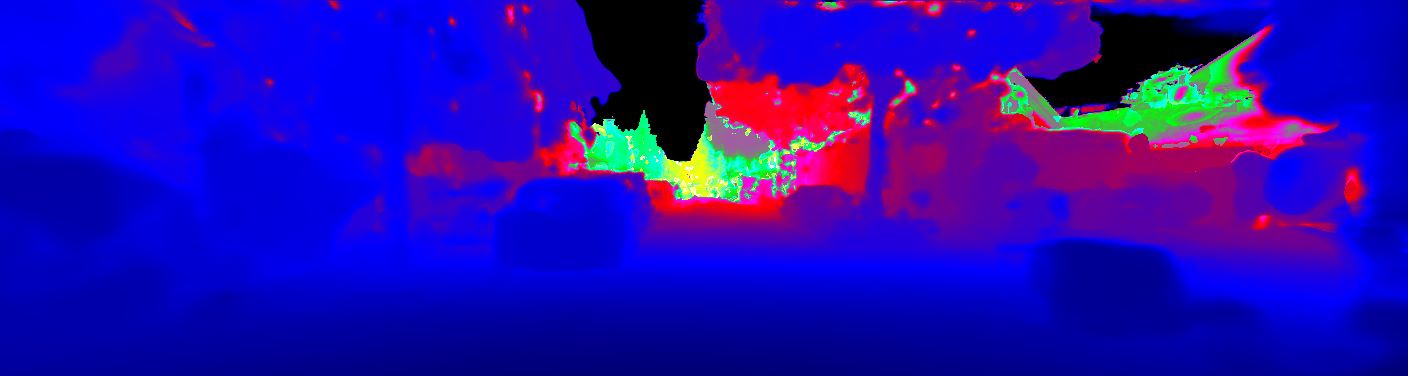}
    \end{tabular}
    \caption{{\bf Limitations in Drop90 Feed-Foward Inference}. Our performance degenerates in regions with insufficient overlap between reference and target frames, e.g., the bottom of the images.}
    \label{fig: failure case}
\end{figure}

Although our approach achieves superior performance on NVS from sparse urban images, it still suffers degeneration when there is less overlap between the target image and reference images. In this case, 3D sample points will not query accurate appearance information from nearby images, leading to blurry rendering at the bottom of the test images, as illustrated in \cref{fig: failure case}. This problem may be mitigated in future work by improving the 3D CNN's capacity to directly predict high-frequency appearance instead of querying color from reference images.

\end{document}